\documentclass[letterpaper, 10 pt, journal, twoside]{IEEEtran}
\usepackage{comment}
\usepackage{siunitx}
\usepackage{relsize}
\usepackage{ifthen}
\usepackage[colorinlistoftodos]{todonotes}

\usepackage{graphics} %
\usepackage{rotating}
\usepackage{color}
\usepackage{enumerate}
\usepackage[T1]{fontenc}
\usepackage{psfrag}
\usepackage{epsfig} %
\usepackage{booktabs}
\usepackage{graphicx,url}
\usepackage{multirow}
\usepackage{array}
\usepackage{latexsym}
\usepackage{amsfonts}
\usepackage{amsmath}
\usepackage{amssymb}
\usepackage{xstring}
\usepackage{multirow}
\usepackage{xcolor}
\usepackage{prettyref}
\usepackage{flexisym}
\usepackage{bigdelim}
\usepackage{breqn} %
\usepackage{listings}
\let\labelindent\relax
\usepackage{enumitem}
\usepackage{xspace}
\usepackage{bm}
\graphicspath{{./figures/}}
\usepackage{tikz}
\usetikzlibrary{matrix,calc}

\usepackage{mdwlist}

\let\itemize\undefined
\makecompactlist{itemize}{stditemize}

\usepackage{etex}
\usepackage{algorithm, algorithmic}
\usepackage{subcaption}
\usepackage{amsmath}

\usepackage{tabularx, colortbl}

\usepackage{filecontents}

\DeclareCaptionLabelSeparator{periodspace}{.\quad}
\IEEEoverridecommandlockouts

\usepackage[numbers,sort&compress]{natbib}
\usepackage{multicol}
\usepackage[bookmarks=true]{hyperref}
\usepackage{amsthm}
\newtheoremstyle{mystyle}%
  {}%
  {}%
  {\itshape}%
  {}%
  {\bfseries}%
  {.}%
  { }%
  {\thmname{#1}\thmnumber{ #2}\thmnote{ (#3)}}%
\theoremstyle{mystyle}
\newrefformat{prob}{Problem\,\ref{#1}}
\newrefformat{def}{Definition\,\ref{#1}}
\newrefformat{sec}{Section\,\ref{#1}}
\newrefformat{sub}{Section\,\ref{#1}}
\newrefformat{prop}{Proposition\,\ref{#1}}
\newrefformat{app}{Appendix\,\ref{#1}}
\newrefformat{alg}{Algorithm\,\ref{#1}}
\newrefformat{cor}{Corollary\,\ref{#1}}
\newrefformat{thm}{Theorem\,\ref{#1}}
\newrefformat{lem}{Lemma\,\ref{#1}}
\newrefformat{fig}{Fig.\,\ref{#1}}
\newrefformat{tab}{Table\,\ref{#1}}

\newcommand{\cf}{\emph{cf.}\xspace}

\newcommand{\bdmath}{\begin{dmath}}
\newcommand{\edmath}{\end{dmath}}
\newcommand{\beq}{\begin{equation}}
\newcommand{\eeq}{\end{equation}}
\newcommand{\bdm}{\begin{displaymath}}
\newcommand{\edm}{\end{displaymath}}
\newcommand{\bea}{\begin{eqnarray}}
\newcommand{\eea}{\end{eqnarray}}
\newcommand{\beal}{\beq \begin{array}{ll}}
\newcommand{\eeal}{\end{array} \eeq}
\newcommand{\beas}{\begin{eqnarray*}}
\newcommand{\eeas}{\end{eqnarray*}}
\newcommand{\ba}{\begin{array}}
\newcommand{\ea}{\end{array}}
\newcommand{\bit}{\begin{itemize}}
\newcommand{\eit}{\end{itemize}}
\newcommand{\ben}{\begin{enumerate}}
\newcommand{\een}{\end{enumerate}}

\newcommand{\calC}{{\cal C}}

\newcommand{\setal}{~\emph{et~al.}\xspace}
\newcommand{\eg}{\emph{e.g.,}\xspace}
\newcommand{\ie}{\emph{i.e.,}\xspace}
\newcommand{\myParagraph}[1]{{\bf #1.}\xspace}

\newcommand{\hide}[1]{}

\newcommand{\hiddenText}{{\color{gray} hidden text.}}
\newcommand{\hideWithText}[1]{\hiddenText}

\newcommand{\tran}{^{\mathsf{T}}}

\newcommand{\Real}[1]{ { {\mathbb R}^{#1} } }

\newcommand{\vf}{\boldsymbol{f}}

\newcommand{\vtheta}{\boldsymbol{\theta}}

\newcommand{\blue}[1]{{\color{blue}#1}}

\newcommand{\linkToPdf}[1]{\href{#1}{\blue{(pdf)}}}
\newcommand{\linkToPpt}[1]{\href{#1}{\blue{(ppt)}}}
\newcommand{\linkToCode}[1]{\href{#1}{\blue{(code)}}}
\newcommand{\linkToWeb}[1]{\href{#1}{\blue{(web)}}}
\newcommand{\linkToVideo}[1]{\href{#1}{\blue{(video)}}}
\newcommand{\linkToMedia}[1]{\href{#1}{\blue{(media)}}}
\newcommand{\award}[1]{\xspace} %

\usepackage{float}
\usepackage{hyperref}
\usepackage{graphicx}
\usepackage{epstopdf}
\usepackage{epsfig}
\usepackage{amsmath}
\usepackage[capitalise]{cleveref}
\usepackage[normalem]{ulem}

\DeclareFontFamily{U}{stix2bb}{}
\DeclareFontShape{U}{stix2bb}{m}{n} {<-> stix2-mathbb}{}

\newcommand{\myparagraph}[1]{{\bf #1.}}

\newcommand{\name}{Clio\xspace}
\newcommand{\batch}{batch\xspace}

\newcommand{\open}{open-set\xspace}

\newcommand{\OPEN}{Open-Set\xspace}
\newcommand{\closed}{closed-set\xspace}

\newcommand{\cg}{ConceptGraphs\xspace}
\newcommand{\thres}{task\xspace}

\newcommand{\probP}{\text{I\kern-0.15em P}}

\newcommand{\ib}{IB\xspace}
\newcommand{\pygivenx}{p({y}|{x})}
\renewcommand{\vf}{f}
\renewcommand{\vtheta}{\theta}

\renewcommand{\baselinestretch}{0.931}

\DeclareRobustCommand{\IEEEauthorrefmark}[1]{\smash{\textsuperscript{\footnotesize #1}}}

\newcommand{\crossout}[1]{}
\newcommand{\reda}[1]{#1}  %

\newcolumntype{C}{>{\centering\arraybackslash}X} 

\usepackage{xr}

\title{\vspace{0.25in}\LARGE \bf{ \emph{\name:} Real-time Task-Driven Open-Set 3D Scene Graphs}}

\author{Dominic Maggio$^*$\IEEEauthorrefmark{1},
                  Yun Chang$^*$\IEEEauthorrefmark{1},
                  Nathan Hughes$^*$\IEEEauthorrefmark{1},
                  Matthew Trang\IEEEauthorrefmark{2},\\
                  Dan Griffith\IEEEauthorrefmark{2},
                  Carlyn Dougherty\IEEEauthorrefmark{2},
                  Eric Cristofalo\IEEEauthorrefmark{2},
                  Lukas Schmid\IEEEauthorrefmark{1},
                  Luca Carlone\IEEEauthorrefmark{1}
    \thanks{Manuscript received: April 24, 2024; Accepted August 10, 2024.
    This letter was recommended for publication by Editor S. Behnke upon evaluation of the Associate Editor and Reviewers' comments.
    This work was supported in part by the NSF Graduate Research Fellowship Program under Grant 2141064, the 
     Swiss National Science Foundation (SNSF) grant No. 214489, 
    MIT Lincoln Laboratory's \textit{Autonomy al Fresco} program, 
    the ARL DCIST program, and the ONR RAPID program.}
    \thanks{$^1$Laboratory for Information \& Decision Systems, Massachusetts Institute of Technology 
    Cambridge, MA, USA. Email: \{drmaggio, yunchang, na26933, lschmid, lcarlone\}@mit.edu.}
    \thanks{$^2$MIT Lincoln Laboratory, Lexington, MA, USA. 
    Email: \{matthew.trang, dan.griffith, eric.cristofalo, carlyn.dougherty\}@ll.mit.edu.}
    \thanks{$^*$equal contribution.}
    \thanks{Digital Object Identifier (DOI): see top of this page.}
    \thanks{\tiny DISTRIBUTION STATEMENT A. Approved for public release. Distribution is unlimited.
    This material is based upon work supported by the Under Secretary of Defense for Research and Engineering under Air Force Contract No. FA8702-15-D-0001. Any opinions, findings, conclusions or recommendations expressed in this material are those of the author(s) and do not necessarily reflect the views of the Under Secretary of Defense for Research and Engineering.
    © 2024 Massachusetts Institute of Technology.
    Delivered to the U.S. Government with Unlimited Rights, as defined in DFARS Part 252.227-7013 or 7014 (Feb 2014). Notwithstanding any copyright notice, U.S. Government rights in this work are defined by DFARS 252.227-7013 or DFARS 252.227-7014 as detailed above. Use of this work other than as specifically authorized by the U.S. Government may violate any copyrights that exist in this work.}
}

\begin{document}
\bstctlcite{IEEEexample:BSTcontrol}

\begin{minipage}{\textwidth} \copyright 2024 IEEE. Personal use of this material is permitted. Permission from IEEE must be obtained for all other uses, in any current or future media, including reprinting/republishing this material for advertising or promotional purposes, creating new collective works, for resale or redistribution to servers or lists, or reuse of any copyrighted component of this work in other works.\\ Please cite this paper as:\\
  \begin{verbatim} 
    @ARTICLE{Maggio2024Clio,
    title={Clio: Real-time Task-Driven Open-Set 3D Scene Graphs}, 
    author={Maggio, Dominic and Chang, Yun and Hughes, Nathan and Trang, Matthew and 
    Griffith, Dan and Dougherty, Carlyn and Cristofalo, Eric and 
    Schmid, Lukas and Carlone, Luca},
    journal={IEEE Robotics and Automation Letters}, 
    year={2024},
    volume={9},
    number={10},
    pages={8921-8928},
    doi={10.1109/LRA.2024.3451395}
  } \end{verbatim} 
\end{minipage}

\markboth{IEEE Robotics and Automation Letters. Preprint Version. Accepted August, 2024}
{Maggio \MakeLowercase{\textit{et al.}}: Clio: Real-time Task-Driven Open-Set 3D Scene Graphs} 

\maketitle

\begin{abstract}
Modern tools for class-agnostic image segmentation (\eg SegmentAnything) and \open semantic understanding (\eg CLIP) provide unprecedented opportunities for robot perception and mapping.
While traditional \closed metric-semantic maps were restricted to tens or hundreds of semantic classes, we can now build maps with a plethora of objects and countless semantic variations.
This leaves us with a fundamental question: \emph{what is the right granularity for the objects (and, more generally, for the  semantic concepts) the robot has to include in its map representation?} While related work implicitly chooses a level of granularity by tuning thresholds for object detection, %
we argue that such a choice is intrinsically {task-dependent}.
The first contribution of this paper is to propose a \emph{task-driven 3D scene understanding} problem,   
where the robot is given a
list of tasks\crossout{,specified}
in natural language, and has to 
 select the granularity and the subset of objects and scene structure to retain in its map that is sufficient to complete the tasks. We show that this problem can be naturally formulated using the \emph{Information Bottleneck} (IB), an established information-theoretic framework to discuss task-relevance. 
The second contribution is an algorithm for task-driven 3D scene understanding based on an \emph{Agglomerative IB} approach, that is able to cluster 3D primitives in the environment into task-relevant objects and regions.\crossout{, and executes incrementally.}
The third contribution is to integrate our task-driven clustering %
algorithm into a real-time pipeline, named \emph{\name}, that constructs a hierarchical 3D scene graph of the environment online and using only onboard compute.
\crossout{, as the robot explores it.} 
Our final contribution is an extensive experimental campaign showing that
\name not only allows real-time construction of compact \open 3D scene graphs, but also improves the accuracy of task 
execution by limiting the map to relevant semantic concepts. 

\end{abstract}
 
\begin{IEEEkeywords}
  Mapping, Deep Learning for Visual Perception, Semantic Scene Understanding
\end{IEEEkeywords}

\IEEEpeerreviewmaketitle

\section{Introduction}
\label{sec:intro}

\begin{figure}[hbt]
    \vspace{1.5mm}
    \centering
    \includegraphics[width=0.5\textwidth]{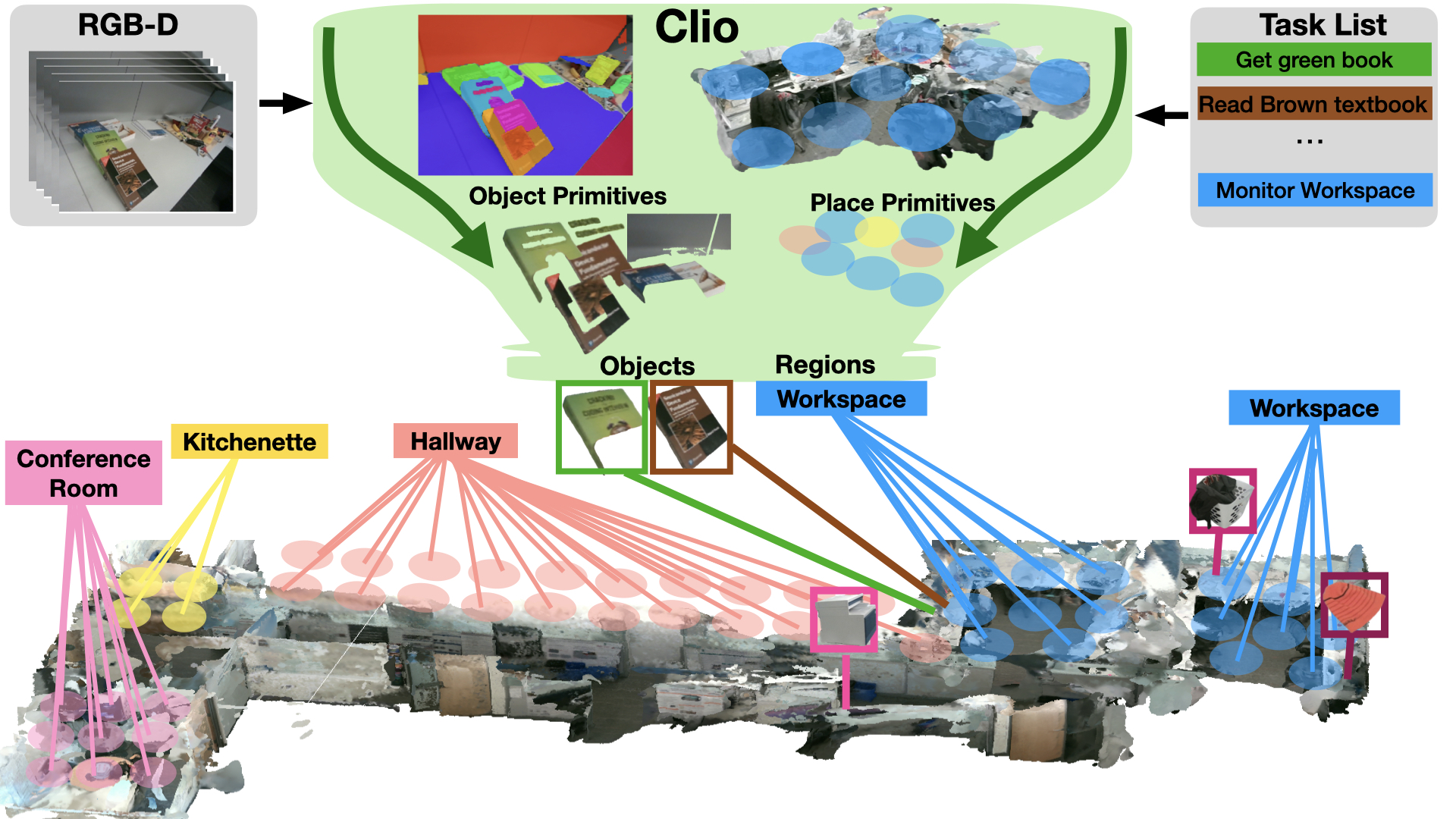} \\
    \caption{%
        We propose \emph{\name{}}, a novel approach for building task-driven 3D scene graphs in real-time with embedded open-set semantics.
        We draw inspiration from the classical Information Bottleneck 
        principle to form task-relevant clusters of object primitives %
        given a set of natural 
        language tasks ---such as ''Read brown textbook''---
        and by clustering the scene into task-relevant semantic regions such as ``Kitchenette'' or ``Workspace''. 
    }\label{fig:main_fig}
\end{figure}
 
\IEEEPARstart{A} fundamental problem in robotics is to create a useful map representation 
of the scene observed by the robot, where usefulness is measured by the ability of the robot to use the map to complete 
tasks of interest~\cite{Soatto16iclr-visualRepresentations,Cadena16tro-SLAMsurvey}. 
Recent works, including~\cite{Armeni19iccv-3DsceneGraphs, Rosinol20rss-dynamicSceneGraphs, Wu21cvpr-SceneGraphFusion, Hughes22rss-hydra,Hughes24ijrr-hydraFoundations}, build metric-semantic 3D maps by 
detecting objects and regions corresponding to a closed set of semantic labels.
However, \closed detection is inherently limited in terms of the set of concepts that can be represented and does not cope well with the intrinsic ambiguity and variability of natural language.
In order to overcome these limitations, 
a new set of approaches~\cite{Jatavallabhula23rss-ConceptFusion, Gu24icra-conceptgraphs} has begun to leverage vision-language foundation models for \open semantic understanding. 
These approaches use a class-agnostic 
segmentation network~\cite{Kirillov23iccv-SegmentAnything} (SegmentAnything or SAM) 
to generate fine-grained segments of the image and then apply a foundation model~\cite{Radford21icml-clip} to get an embedding vector describing the \open semantics of each segment. 
Objects are then constructed by associating segments whenever their embedding vectors are within a 
predefined similarity threshold. %
These approaches, however, leave to the user the difficult task of tuning suitable thresholds to control the number of segments that are extracted from the scene as well as the threshold used to decide whether two segments have to be clustered together. 
More importantly, these methods 
 do not capture intuition that the choice of semantic concepts in the map is not just driven by semantic similarity, but it is intrinsically \emph{task-dependent}. 
For example, consider a robot 
tasked with moving a piano across a room. The robot gains almost no value by distinguishing the 
location of all the keys and strings, but can instead complete the task by considering 
the piano as one large object. On the other hand, a robot tasked with playing
the piano must consider the piano as many objects (i.e., the keys). A robot tasked with tuning the 
piano must view the piano as even more objects --- considering the strings, tuning pins, and so forth.
Likewise, questions such as if a pile of clothes should be represented 
as a single pile or as individual clothes, or if 
a forest should be represented as single area of landscape or as branches, leaves, trunks, etc., remains ill-posed until
we specify the tasks that the representation has to support.
Humans not only take into account the task when (consciously or unconsciously) deciding which objects to represent and how, but
are also able to consequently ignore parts of a scene 
that are irrelevant to the task~\cite{Treisman80cp-PsychologyAttention}. %

\myParagraph{Contributions} 
Our first contribution (Section~\ref{sec:problem}) is to state the \emph{task-driven 3D scene understanding problem}, 
where the robot is given a list of tasks, specified in natural language, and is required to build a minimal map representation
 that is sufficient to complete the given tasks. 
More specifically, we assume the robot is capable of perceiving task-agnostic primitives in the environment, in the form of a large set of 3D object segments and 3D obstacle-free places, and has to cluster them into a task-relevant compressed representation which only contains relevant objects and regions (\eg rooms). 
This problem can be naturally formulated using the classical \emph{Information Bottleneck} (IB)~\cite{Tishby01accc-IB} theory, which 
also provides algorithmic approaches for task-driven clustering. 

Our second contribution (Section~\ref{sec:obj_clustering}) is to apply the Agglomerative IB algorithm from~\cite{Slonim99nips-AgglomerativeIB} to the problem of task-driven 3D scene understanding. In particular, we show how to obtain the probability densities required by the algorithm in~\cite{Slonim99nips-AgglomerativeIB} using CLIP embeddings, and show that  the resulting algorithm can be executed incrementally as the robot explores the environment, with a computational complexity that does not increase with the environment size.

\begin{figure}[t]
    \vspace{1.5mm}
    \centering
    \includegraphics[width=0.489\textwidth]{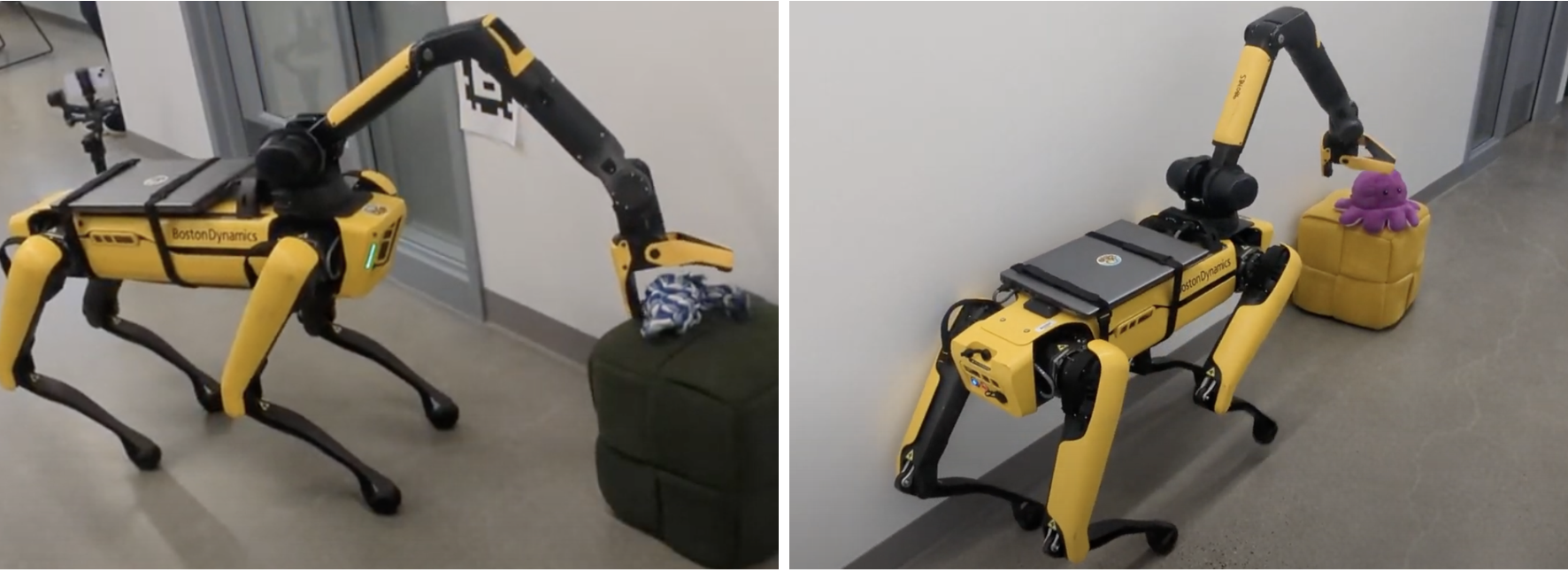}\vspace{-2mm}\\
    \caption{%
        \name generates a 3D scene graph in real-time using a laptop carried by Spot. 
        We show that Spot is able to execute grasping commands, expressed in natural language, 
        using \name's task-driven 3D scene graph.
    }\label{fig:spot}
\end{figure}

Our third contribution (Section~\ref{sec:map}) is to include the proposed task-driven clustering algorithm into a real-time system, named \emph{\name} (Fig.~\ref{fig:main_fig}).
\name takes a list of tasks specified in natural language at the beginning of operation: for instance, these can be the tasks the robot is 
envisioned to perform during its lifetime or during its current deployment.
Then, as the robot operates, \name creates a hierarchical map, namely a \emph{3D scene graph}, of the environment in real-time, 
where the representation only retains task-relevant objects and regions.  
Contrary to current approaches for \open 3D scene graph construction (\eg~\cite{Gu24icra-conceptgraphs}) which are restricted to off-line operation when 
querying large vision-language models (VLMs)~\cite{Liu23neurips-llava} and Large Language Models (LLMs) such as~\cite{OpenAI23arxiv-GPT4}, \name runs 
in real-time and onboard and only relies on lightweight foundation models, such as CLIP~\cite{Radford21icml-clip}.

We demonstrate \name on the Replica dataset~\cite{Straub19arxiv-replica} and in four real environments (Section~\ref{sec:experiments}) --- an apartment, 
an office, a cubicle, and a large-scale building scene. 
We also show real-time onboard mapping with \name on a 
Boston Dynamics Spot quadruped with a robotic arm~(\cref{fig:spot}). %
\name not only allows real-time \open 3D scene graph construction, but also improves the accuracy of task 
execution by limiting the map to relevant objects and regions.
We release \name open-source at \url{https://github.com/MIT-SPARK/Clio} along with our custom datasets.

\section{Related Work}
\label{sec:related_works}

\myParagraph{Foundation Models in Robotics and Vision}
The recent emergence of  
vision-language models~\cite{Radford21icml-clip, Oquab23arxiv-dinov2, Liu23neurips-llava} and large 
language models~\cite{OpenAI23arxiv-GPT4} has led to numerous works exploring their potential for 
3D scene understanding~\cite{Hong23neurips-3dllm, Zhao23iccv-textpsg}
and robot planning~\cite{Chang23arxiv-goat, Garg23corl-robohop, 
Huang23icra-languageMaps}. 
Multiple works have surveyed the state of the art in 
foundation models along with their limitations~\cite{Firoozi23arxiv-FoundationModelSurvey, 
Sharma24cvpr-visionCheckUp, Tong24cvpr-eyesWideShut}.
Class-agnostic segmentation networks~\cite{Kirillov23iccv-SegmentAnything, Zhao23Arxiv-FastSam} 
have been coupled with foundation models to enable 
open-set image segmentation~\reda{\cite{Zhou23cvpr-zegclip, 
Minderer22eccv-owlVit, Liu23arxiv-groundingDino, 
Li22iclr-lseg, Ding23icml-maskclip, Cheng21cvpr-mask2former}}.
Recent works have also explored direct class-agnostic 3D segmentation~\cite{Huang23arxiv-segment3d}.
Saliency detection 
has been used to identify parts of an image that a human would 
likely notice first~\cite{Roberts12spie}. Here, instead of visual 
saliency, we desire to create task-driven maps of a scene.

\myParagraph{Foundation Models for 3D Mapping}
Recent work has coupled foundation models with neural radiance fields~\cite{Mildenhall21acm-nerf} and 
Gaussian Splatting~\cite{Kerbl23Ttog-GaussianSplatting}. Kerr\setal~\cite{Kerr23iccv-lerf} propose LERF, 
which constructs a radiance 
field that can render dense CLIP vectors of the scene. LERF can be queried via text and estimate which parts of 
the scene are most similar to the query using an augmented cosine similarity score. Qin\setal~\cite{Qin24cvpr-langsplat} 
develop LangSplat which builds upon LERF by using Gaussian Splatting to 
create a 3D scene language map with a substantial speedup.
Blomqvist\setal~\cite{Blomqvist23iros-neuralFeatureFields} develop an approach to 
incrementally construct a neural semantic map for SLAM. Kim\setal~\cite{Kim24cvpr-garfield} construct a 
hierarchical neural map that renders at different levels of granularity, clustering and dividing objects into parts.
Taioli\setal~\cite{Taioli23iccv-languageRnrMap} use CLIP to construct an implicit grid map that can be 
queried via text. %

Several works incorporate \open detection into 3D maps of a 
scene~\cite{Peng23cvpr-OpenScene3D, Ha22corl-openWorld, Wang23ral-SeMLaPSRS, Koch243dv-lang3dsg, Yamazaki24icra-openFusion, Kassab24icra-lexis}.
Chang\setal~\cite{Chang23corl-ovsg} perform open-vocabulary mapping combined with a graph neural network
trained on a closed set to map objects and their relationships.
Takmaz\setal~\cite{Takmaz23neurips-openmask3d} develop a method for open-set instance segmentation.
Jatavallabhula\setal~\cite{Jatavallabhula23rss-ConceptFusion} generate a semantic 
3D point cloud where CLIP vectors are assigned to each point. 
Most similar to ours is \cg~\cite{Gu24icra-conceptgraphs}, which constructs a 3D graph of objects with 
edges connecting objects via their relationships as assigned 
with an LLM~\cite{OpenAI23arxiv-GPT4}. \cg uses CLIP and SAM to cluster 
a scene into objects defined by their semantic and geometric similarity to each other. Optionally, \cg queries 
a large vision-language model~\cite{Liu23neurips-llava} using multiple views of each object to 
compute a succinct description of the object.
Objects can be then queried either with cosine similarity via CLIP or 
with the LLM. \reda{Concurrently, Werby\setal~\cite{Werby24rss-hovsg} demonstrate large-scale \open semantics using a 
hierarchical 3D scene graph}, but does not run in realtime.

\myParagraph{Task-Driven Representations}
The classical Information Bottleneck~\cite{Tishby01accc-IB} aims to compress a  given signal while 
preserving the mutual information between the compressed representation and another signal of interest. The initial work~\cite{Tishby01accc-IB} 
has been extended into a bottom-up clustering method known as the Agglomerative IB~\cite{Slonim99nips-AgglomerativeIB}. 
We build on IB theory with the goal of compressing a scene representation into 
clusters of relevant objects and regions for a 
given set of tasks.
 Gordon\setal~\cite{Gordon03iccv-imageIB} extend the Information Bottleneck to compress a set of individual images into clusters such that each cluster 
preserves information about the context of the images contained in the cluster. 
Wang\setal~\cite{Wang23neurips-clipIB} use IB 
for attribution between image 
and text inputs of VLMs with experiments performed with CLIP. 
\reda{Larsson\setal~\cite{Larsson21ral-infoTheoreticAbstractionsPlanning, Larsson20tro-qTree} 
leverage the Agglomerative IB to obtain an optimal occupancy map compression for agents with limited resources.}

Soatto and Chiuso~\cite{Soatto16iclr-visualRepresentations} derive expressions for minimally sufficient 
scene representations that preserve 
 relevant information about some task of interest, 
and~\cite{Parameshwara23arxiv-visualFoundationModels} develops theory around 
constructing foundation models of physical scenes. 
Eftekhar\setal~\cite{Eftekhar24iclr-bottneckedTaskRepresentations} compress visual observations in a task-relevant manner. Their work uses a learned 
codebook module that takes in a current agent's action along with the task and sensor data, and outputs an 
action to step towards the goal for navigation.
Another line of work detects regions of interest in images based on affordances~\cite{Mur23icra-bayesianAffordance} 
and creates 3D maps of affordances of objects in a scene~\cite{Mur-Labadia23iccv-affordanceMapping}.

\section{Problem Formulation: \\ Task-Aware 3D Scene Understanding}
\label{sec:problem}
While many researchers would agree that a map representation has to be task-dependent, to date there is 
no general framework to establish what is the right granularity for the semantic concepts included in the robots' metric-semantic 3D map.
 This gap has been partially motivated by the difficulty of providing rich task descriptions, with the result that existing 
 task-driven representation frameworks in vision and robotics are either too narrow or too computationally expensive~\cite{Soatto14iclr}.

In this paper, we leverage two key insights. First of all, progress in vision-language models has brought together visual information and text descriptions in a way that was not possible before. This greatly simplifies the problem of task description: we can just 
state the task as a list of language instructions the robot is expected to execute during its lifetime or during its current deployment (\eg ``wash the dishes'', ``fold the clothes'', ``pick up toys and place them on the shelves'') and use VLMs to relate these instructions to visual data. Below, we denote the list of tasks with the symbol $Y$.
Second, modern foundation models for task-agnostic segmentation provide a way to over-segment an image into a potentially large number of segments, which we can reproject to 3D. Similarly, using geometric segmentation techniques, we can easily segment environments into a large number of obstacle-free places~\cite{Hughes22rss-hydra}. In the following, we refer to the task-agnostic 3D segments and places as \emph{task-agnostic primitives} and denote them with $X$; intuitively, these provide a superset of the concepts we want to retain in our map. 

Using these insights we formulate task-aware 3D scene understanding as the problem of compressing the task-agnostic primitives $X$ into a cluster of task-relevant concepts $\tilde{X}$, which are maximally informative about the tasks $Y$. 
This naturally leads to the Information Bottleneck principle.

\myParagraph{Task-Aware 3D Scene Understanding as an Information Bottleneck}
Similar to the setup of the well-known Information Bottleneck (\ib)~\cite{Tishby01accc-IB},
we have an original signal $X$ (\ie the set of task-agnostic primitives),
which provides some information about the signal $Y$ (\ie the list of tasks).
Our goal is to find a more compact signal $\tilde{X}$ ---representing the task-relevant concepts--- 
that compresses $X$ while retaining task-relevant information. 
Mathematically, we are going to define the task-relevant clusters $\tilde{X}$ using the probability 
distribution $p(\tilde{x}|x)$, which represents the probability that a task-agnostic primitive in $x$ belongs to cluster in $\tilde{x}$.
\ib formulates the computation of the task-relevant clusters $\tilde{X}$ (or, equivalently, the probability $p(\tilde{x}|x)$) as the solution of the following optimization: 
\begin{equation}
	\label{eq:ib}
	\textstyle\min_{p(\tilde{x}|x)} I(X;\tilde{X}) - \beta I(\tilde{X}; Y),
\end{equation}
where $I(\cdot;\cdot)$ denotes the mutual information between two random variables.
Intuitively, problem~\eqref{eq:ib} compresses $X$ by minimizing the mutual information between the
original signal $X$ and compressed signal $\tilde{X}$, while rewarding the task-relevance of the compressed representation 
through the mutual information between the compressed signal $\tilde{X}$ and the task $Y$.
The parameter $\beta$ controls the desired balance between the two terms (\ie the amount of compression).

The result of~\eqref{eq:ib} is a set of clusters: intuitively, these clusters group 3D segments into objects and 3D places into regions (\eg rooms) at the right granularity, as required by the task.
\crossout{Problem~\eqref{eq:ib} can be solved by an iterative algorithm with proven convergence~\cite{Tishby01accc-IB}.}
Below, 
we discuss algorithms that can better take advantage of the structure of our problem and shed light on how to compute the 
 distributions and mutual information terms arising in~\eqref{eq:ib} in practice.

\section{Task-Driven Clustering}
\label{sec:obj_clustering}

In our problem, the task-agnostic primitives have geometric attributes, which provide 
a strong inductive bias for our clustering (\ie we might want to merge together nearby segments, and avoid merging segments that are far away). 
To enforce this inductive bias, we consider and extend the Agglomerative IB approach of~\cite{Slonim99nips-AgglomerativeIB}, which forms task-relevant clustering by iteratively merging neighboring primitives.
\reda{In this section, we first provide relevant background on the Agglomerative IB, then present an incremental version  of the Agglomerative IB algorithm to support 
real-time mapping, and lastly tailor the IB formulation to the use of \open vision-language features for task-aware scene understanding}.

\myParagraph{Agglomerative Information Bottleneck}
The Agglomerative IB method is
a bottom-up merging approach to solving the \ib problem~\cite{Slonim99nips-AgglomerativeIB}.
The method initializes the task-relevant clusters $\tilde{X}$ to the task-agnostic primitives $X$; then, at each iteration, it merges adjacent clusters using a task-driven metric.
In particular, it computes 
a weight $d_{ij}$ for each possible merge between \emph{adjacent} clusters $\tilde{x}_i$ and $\tilde{x}_j$ as:
\begin{equation}\label{eq:distortion}
	d_{ij} = (p(\tilde{x}_i) + p(\tilde{x}_j)) \cdot D_{\textrm{JS}}[p(y|\tilde{x}_i), p(y|\tilde{x}_j)],
\end{equation}
where $D_{\textrm{JS}}$ is the Jensen-Shannon divergence.
Intuitively, the weight $d_{ij}$ is a measure of the dissimilarity of the probability distributions of the two clusters.
In particular, the algorithm iteratively merges the clusters corresponding to the smallest weight,
thus solving \ib in a greedy manner. 
The process can be understood as iteratively merging nearby nodes in a graph, where the graph edges represent allowable merges.

{As suggested in~\cite{Slonim99nips-AgglomerativeIB}, at each iteration $k$, we also compute} 
\begin{equation}\label{eq:delta}
	\delta(k) = \frac{I(\tilde{X}_k;Y) - I(\tilde{X}_{k-1} ; Y)}{I(X;Y)} 
\end{equation}
as a measure of the fractional loss of information corresponding to a merge operation, 
and terminate the algorithm when $\delta(k)$ exceeds a threshold $\bar{\delta}$. 
$\bar{\delta}$ regulates the amount of compression where a value of 0 returns %
the original set of primitives and a value of 1 returns fully merged primitives, %
playing a similar role as  the parameter $\beta$ in eq.~\eqref{eq:ib}.
The pseudocode of the algorithm in given in \cref{sec:aib}.

\myParagraph{Incremental Agglomerative \ib}
In our problem, we expect the map to grow over time, hence it is paramount to bound the computational complexity of the 
Agglomerative IB. Towards this goal, we propose an incremental version of the algorithm that can be executed online as the robot explores the environment.
Our key observation is that if the graph of primitives in input to the algorithm has multiple connected components (\eg 3D object segments in different rooms), then the clustering can we performed independently on each connected component (intuitively, there are no edges, hence no potential merges, between different components). 
Moreover, it is easy to show that the variable $\delta(k)$ in~\eqref{eq:delta} (used in the stopping condition of the algorithm) can be computed independently 
for each connected component, and does not need to be recomputed for connected components that are not affected by new measurements.
This allows the robot to cluster incrementally while supporting a real-time stream of new primitives as it maps the environment. We report the pseudocode of our incremental algorithm in \cref{sec:iaib_proof}, while next we discuss how to set the required distributions.

\myParagraph{Task-Relevant Conditional Distributions}
The Agglomerative \ib algorithm requires defining the conditional probability $\pygivenx$, which can be understood as the task-relevance of
each primitive.
We use CLIP~\cite{Radford21icml-clip} to produce an embedding $\vf_{x_i}$ for each primitive $x_i \in X$
and an embedding $\vf_{t_j}$ for each task $t_j \in Y$.
For each primitive $x_i$, we compute its cosine similarity score $\phi(\vf_{x_i},\vf_{t_j})$ to all task embeddings.
We further add a \emph{null} task $t_0$ and assign it a score $\alpha$, which is chosen as a lower-bound on the 
cosine similarity under which a primitive is not relevant for any of the given tasks.

We perform a pre-pruning step on primitives that have the highest similarity with the null task,
for which we set $p(y|x_i)$ to be a one-hot vector with a probability of 1 on the null task.
Furthermore, to emphasize the ranking of task similarities,
we set all task similarities that are not in the top $k$ most similar tasks to 0
and multiply the top $l$ task by $k - l + 1$.
Formally,  given $m$ tasks, we first define $\vtheta(x_i) \in \Real{m+1}$: %
\begin{equation}
\label{eq:pyx_vanilla}
	\vtheta(x_i)_j =
	\begin{cases}
      \alpha, & \text{if}\ j = 0 \\
      \phi(\vf_{xi},\vf_{tj}), &\text{if}\ j = 1,\ldots,m %
    \end{cases}
\end{equation}
and then write $\pygivenx$ in terms of $\vtheta$ as,
\begin{equation}
\hspace{-3mm}\label{eq:pyx}
	p(y|x_i) =
	\begin{cases}
		[1\;0\;\ldots\;0]\tran, & \text{if} \  \max_{t_j} \phi(\vf_{xi},\vf_{tj}) \!\!<\!\! \alpha \\
		\eta\sum_{l=1}^k \gamma_l(\vtheta(x_i)), & \text{otherwise}
    \end{cases}
\end{equation}
where $\eta$ is a normalization constant and {$\gamma_l$ preserves only the top $l$ values while setting all others to $0$.}
This choice of $\pygivenx$ effectively assigns large values in $\pygivenx$ to the $k$ tasks that have the highest cosine similarity in terms of CLIP embeddings, while also assigning irrelevant primitives to the null task. 
Given this choice of conditional probability, the Agglomerative \ib computes the clusters $\tilde{X}$.

\section{\name: Real-time Task-Driven \\ \OPEN 3D Scene Graphs}
\label{sec:map}

\begin{figure}
    \vspace{1.5mm}
    \includegraphics[width=0.99\columnwidth]{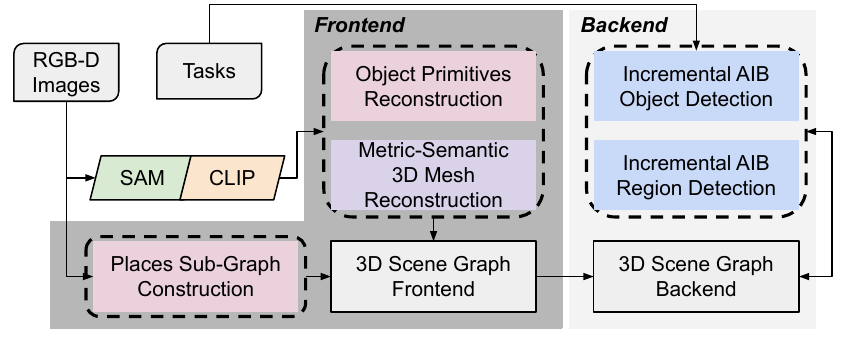}
    \caption{\name's frontend takes in RGB-D sensor data and constructs the graph of object primitives, the graph of places, and the metric-semantic 3D mesh of the background. \name's backend performs Incremental Agglomerative \ib to cluster objects and regions based on a user-specified list of tasks.}
    \label{fig:clio_architecture}
\end{figure}

This section describes \emph{\name}, our real-time system for task-driven \open 3D scene graph construction.
A high-level architecture is shown in Fig.~\ref{fig:clio_architecture}.
\name consists of two main components:
the frontend, where the task-agnostic object and place  primitives are constructed, 
and the backend, where the task-driven object and region clustering is performed.

\subsection{\name Frontend}
\label{sec:frontend}

\myParagraph{3D Object Primitives} We follow the approach of Khronos~\cite{Schmid24rss-khronos} for 3D mesh reconstruction and object primitive extraction. 
Given a live stream of RGB-D images and poses, we run FastSAM~\cite{Zhao23Arxiv-FastSam} and CLIP 
to get semantic segments for each image.
We then temporally associate segments to existing tracks within a temporal window $\tau$.
To enforce consistency, candidate tracks are required to have a cosine similarity above a threshold $\theta_{\text{track}}$\footnote{Note that this threshold is only used to re-identify and track segments over time, while we use our task-driven clustering to group primitives.} and minimum 3D IoU of $\gamma$ with the segment.
Each new segment is then greedily associated to the candidate track with the highest IoU. 
If no association is made, a new track is created.
Finally, if a track has not been associated for $\tau$ seconds, it is terminated.
\crossout{Tracks with a large volume $>\theta_{\text{vol}}$ and with only few observations $<\theta_{\text{obs}}$ are discarded as they are likely background or noise.}
Each track is then reconstructed into a 3D object primitive based on all frames in the track and a final CLIP feature is computed via averaging.
Simultaneously, a coarser reconstruction of the background is performed for every incoming frame.
This approach allows for a dense 3D model to be incrementally constructed with limited computation, while maintaining a high level of detail for the object primitives.

\myParagraph{3D Place Primitives} We follow the approach of Hydra~\cite{Hughes24ijrr-hydraFoundations} to construct the places sub-graph.
We incrementally compute a Generalized Voronoi Diagram of the scene and sparsify it into a graph of places.
To obtain semantic features for the places, we compute a CLIP embedding vector for each input image provided to \name{}.
Each place node is then assigned a feature that is the average of the input CLIP embeddings from all input images that the node centroid is visible.\crossout{from.}
We validate these design choices in \cref{sec:places_exp}.

\subsection{\name Backend}
\label{sec:backend}

\crossout{
After the frontend populates the background mesh, object primitives, and places,
the backend performs clustering to extract and retain task-relevant objects and regions. }

\myparagraph{Task-Driven Object Detection}
\name runs our Agglomerative \ib method on the over-segmented 3D object primitives from the frontend. 
As input to \ib, we construct a graph where the nodes are the object primitives 
and add edges between nodes if the corresponding primitives have 3D bounding boxes 
with non-zero overlap. We compute $\pygivenx$ as described in eq.~\eqref{eq:pyx}. 
In this case, the \emph{null} task can be thought of as background task-irrelevant objects. 
We set $\alpha = 0.23$. %
We provide two versions of \name. The first, \emph{\name-\batch} assumes all primitives 
for the entire scene have first been generated and then clusters all objects segments using eq.~\eqref{eq:delta}. The second, 
\emph{\name-online} takes in a real-time stream of images and constructs a map using our incremental IB algorithm, where 
clustering is only performed again for the connected components affected by the most recent measurements.

\myparagraph{Task-Driven Clustering of Places}
\name 
\crossout{finally}
performs Agglomerative \ib at every backend update to cluster the places primitives nodes into regions,
where each edge in the place graph is considered as a putative merge for clustering.
\crossout{where each place node in the graph is a primitive for IB and each edge in the place graph is considered as a putative merge for clustering.}
We compute $\pygivenx$ between the tasks and place nodes in the same manner as the objects.
\crossout{Each resulting cluster of place nodes is used to create a new region.}
\crossout{Two regions share an edge if any places in the two regions also share an edge.}

\section{Experiments}
\label{sec:experiments}

\begin{figure}
    \vspace{1.5mm}
    \centering
    \begin{subfigure}[b]{0.47\textwidth}
        \includegraphics[width=\textwidth]{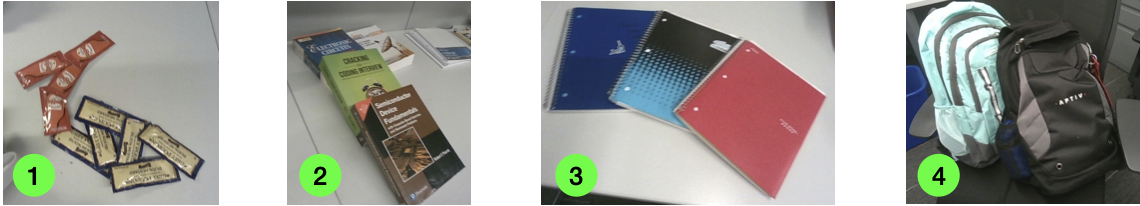}
        \caption{Sample of four regions of the Cubicle dataset}
        \label{fig:scene_parts}
    \end{subfigure}
    \hspace{\fill} %
    \centering
    \begin{subfigure}[b]{0.47\textwidth}
        \includegraphics[width=\textwidth]{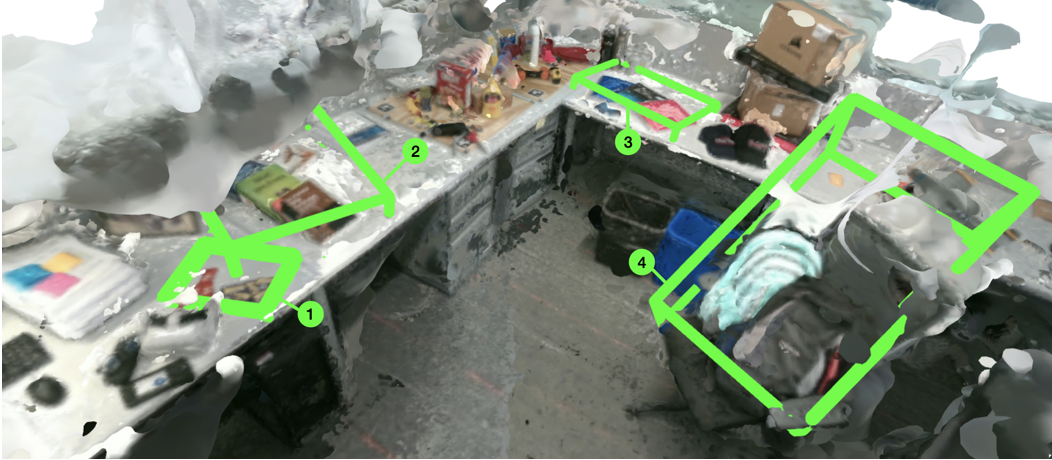}
        \caption{\name clustering results shown for the following tasks: (1) get condiments packets, 
        (2) get textbooks, (3) get notebooks, (4) clean backpacks.}
        \label{fig:ours_course}
    \end{subfigure}
    \hspace{\fill} %
    \begin{subfigure}[b]{0.47\textwidth}
        \includegraphics[width=\textwidth]{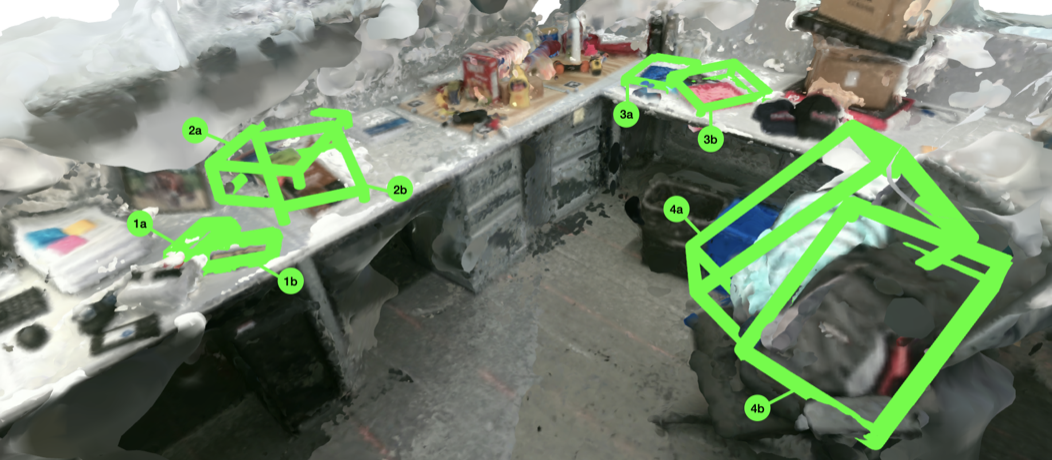}
        \caption{\name clustering results shown for the following tasks: (1a) get hot sauce packets, (1b) get grey 
        poupon packets, (2a) read Cracking the Coding Interview book, (2b) read brown textbook, 
        (3a) pack blue notebooks, (3b) pack red notebook, (4a) get teal backpack, (4b) clean black backpack.}
        \label{fig:ours_fine}
    \end{subfigure}
    \caption{Examples of portions of the Cubicle dataset that require a task to provide rectification 
    of how an object should be defined. The figure showcases Clio's clustering results for two sets of tasks, listed under (b)
 and (c);     %
    14 additional tasks identical for both tests are included in the task list 
    during clustering but not shown for clarity. %
    }
    \label{fig:ib_tasks_visual}
\end{figure}

Our experiments show that Clio (i) constructs more parsimonious and useful map representations (Section~\ref{sec:exp-open-set-objects}), (ii) performs on par with the state of the art in closed-set settings where the task 
is implicitly specified by a closed dictionary (Section~\ref{sec:exp-closed-set-objects}),  (iii) is able to 
cluster the environment into meaningful semantic regions (Section~\ref{sec:places_exp}), and (iv) can support 
task execution on real robots (Section\ref{sec:spot}).

\subsection{\mbox{Open-Set Object Clustering Evaluation}}
\label{sec:exp-open-set-objects}

\myParagraph{Experimental Setup}
To test \name in realistic and diverse scenes, we collect four datasets, in an 
office, an apartment, a cubicle, and a large-scale university building, which covers five floors 
including a machine shop, classroom, lounge, meeting rooms,\crossout{to} 
cluttered workspaces, and an aircraft hangar. 
For the Office, Apartment, and Cubicle
datasets we manually annotate 
ground truth 3D bounding boxes for objects associated to the given set of tasks. 
For evaluation purposes, tasks are chosen such that there is an unambiguous set of objects best suited for 
the tasks, to reduce subjective reasoning over what constitutes a ground truth set of objects. 
A complete list of tasks is provided in \cref{sec:office_labels,sec:apartment_labels,sec:cubicle_labels,sec:builing33_labels}. 
\myParagraph{Metrics}\reda{
Since traditional metrics like precision and recall do not fully capture the performance of open-set object detection, 
we introduce two new metrics: \emph{open-set Recall (osR)} and \emph{open-set Precision (osP)}.
For osR we query the $n$ best objects for every task, where $n$ is the number of ground truth objects relevant for the task,
and report the number of correct detections divided by number of ground truth objects. 
We define osP as the total number of correct detections divided by 
the total number of detections that have at least 90\% cosine similarity score to a task as the most similar object. 
For both metrics, we say a detection is \emph{strict}
if the bounding box of an estimated 
object contains the centroid of the ground truth bounding box, and the bounding box of the 
ground truth object contains the centroid of the estimated bounding box.
We say a detection is \emph{relaxed} if at least one of the two prior conditions is met.
Intuitively, in the worst case, a relaxed detection can be met with an infinitely large estimated bounding box, and a strict detection 
can discount an estimate with meaningful overlap to ground truth. We thus report both criteria. 
We report the F1 score as the harmonic mean of osR and osP and include average IOU of the top $n$ most 
relevant estimated objects, total number of estimated objects (Objs), and average runtime per processed frame (TPF).
}

\myParagraph{Compared Techniques}  As our queries do not include negation or multi-step affordances, we run \cg with only CLIP in place of 
LLava+GPT, as CLIP was shown to have similar performance for these types of queries in~\cite{Gu24icra-conceptgraphs}.
In addition to running \cg and \name, we also test: Khronos, which performs clustering 
as described in~\cite{Schmid24rss-khronos} with
parameters $\theta_{\text{track}} = 0.7$ and $\gamma = 0.4$,\crossout{$\theta_{\text{vol}} = 8.0$
and $\theta_{\text{obs}} = 3$,}
and \name-Prim which only computes the set of input 3D object primitives to \name with
 parameters $\theta_{\text{track}} = 0.9$ and $\gamma = 0.6$\crossout{$\theta_{\text{vol}} = 8.0$ and $\theta_{\text{obs}} = 2$};
essentially, \name-Prim is the output of the \name frontend, hence this comparison allows assessing the effectiveness of the 
IB clustering in \name. 
To show the importance of being task-driven, we further include task-aware versions of the baselines:
Khronos-\thres and \cg-\thres that take the results of Khronos and \cg and remove mapped objects that do not have a high enough ($\alpha = 0.23$) cosine similarity to at least one task in the provided task list.
We include results for both \name-batch, which takes in all primitives of a scene and is executed only once at the end of the mapping session,
and \name-online, which incrementally receives primitives for real-time mapping.
We use CLIP model ViT-L/14 and generate results with an RTX 3090 GPU and Intel \reda{i9-12900K} CPU.
Results are shown in \cref{table:three_d_results}.
Results for OpenCLIP model ViT-H-14 are included 
in \cref{sec:open_clip_objects}.

\begin{table}[h]\scriptsize
    \vspace{1.5mm}
    \setlength{\tabcolsep}{2pt}
    \resizebox{\columnwidth}{!}{%
    {%
    \begin{tabular}{cl ccc ccc ccc}
    \toprule
    &  & \multicolumn{3}{c}{Strict} & \multicolumn{3}{c}{Relaxed} & & & \\
    \cmidrule(r){3-5} \cmidrule(r){6-8}
    Scene & Method & osR$\uparrow$ & osP$\uparrow$ & F1$\uparrow$ & osR$\uparrow$ & osP$\uparrow$ & F1$\uparrow$ & IOU$\uparrow$ & Objs$\downarrow$ & TPF [s]$\downarrow$ \\
    \midrule
    & CG~\cite{Gu24icra-conceptgraphs} & 0.44  & 0.17  & 0.25  & 0.61  & 0.28  & 0.39  & 0.06  & 181 & 2.0  \\ 
    & Khronos~\cite{Schmid24rss-khronos} & 0.78 & 0.12 & 0.21 & 0.83 & 0.11 & 0.20 & 0.17 & 628 & 0.31\\
    & Clio-Prim & 0.72 & 0.09 & 0.16 & 0.72 & 0.10 & 0.17 & \underline{0.18} & 1070 & 0.28\\
    \rowcolor{blue!20} \cellcolor{white} & CG-\thres & 0.44 & \underline{0.38} & 0.41 & 0.61  & \textbf{0.50} & 0.55 & 0.06 & 26 & 2.0 \\
    \rowcolor{blue!20} \cellcolor{white}& Khronos-\thres & 0.78 & 0.14 & 0.24 & 0.83 & 0.14 & 0.24 & 0.17 & 133 & 0.31\\
    \rowcolor{blue!20} \cellcolor{white}& Clio-\batch & \underline{0.83} & 0.33 & \underline{0.47} & \textbf{1.0} & 0.40 & \underline{0.57} & 0.17 & 48 & 0.31$^*$\\
    \rowcolor{blue!20} \cellcolor{white}& Clio-online & \textbf{0.89} & \textbf{0.48} & \textbf{0.62} & \underline{0.89} & \underline{0.48} & \textbf{0.63} & \textbf{0.22} & 92 & 0.30\\
    \midrule
    \multirow{-9}{*}{\rotatebox{90}{Cubicle}}
    & CG~\cite{Gu24icra-conceptgraphs} & 0.24  & 0.09  & 0.13  & 0.52  & 0.16  & 0.25  & 0.07  & 751 & 8.1 \\
    & Khronos~\cite{Schmid24rss-khronos} & \underline{0.67} & 0.24 & 0.35 & 0.67 & 0.25 & 0.36 & \underline{0.15} & 1202 & 0.31\\
    & Clio-Prim & \textbf{0.70} & 0.18 & 0.29 & \underline{0.73} & 0.19 & 0.30 & \textbf{0.17} & 1883 & 0.27\\
    \rowcolor{blue!20}\cellcolor{white}& CG-\thres & 0.19  & 0.37  & 0.25  & 0.45  & \underline{0.63}  & 0.50  & 0.06  & 40 & 8.1 \\
    \rowcolor{blue!20} \cellcolor{white}& Khronos-\thres & 0.55 & 0.28 & 0.37 & 0.55 & 0.30 & 0.38 & 0.12 & 163 & 0.31\\
    \rowcolor{blue!20} \cellcolor{white}& Clio-\batch & 0.64 & \underline{0.45} & \underline{0.53} & \textbf{0.76} & 0.55 & \underline{0.64} & 0.13 & 84 & 0.30$^*$\\
    \rowcolor{blue!20} \cellcolor{white}& Clio-online & 0.55 & \textbf{0.65} & \textbf{0.60} & 0.61 & \textbf{0.69} & \textbf{0.65} & 0.12 & 49 & 0.29\\
    \midrule
    \multirow{-9}{*}{\rotatebox{90}{Office}}
    & CG~\cite{Gu24icra-conceptgraphs}  & 0.38 & 0.17 & 0.23 & 0.62 & 0.25 & 0.35 & 0.07 & 339 & 2.2 \\ 
    & Khronos~\cite{Schmid24rss-khronos} & \underline{0.45} & 0.08 & 0.14 & \textbf{0.76} & 0.12 & 0.21 & \underline{0.11} & 1093  & 0.26\\
    & Clio-Prim & 0.35 & 0.07 & 0.12 & 0.59 & 0.09 & 0.16 & \textbf{0.12} & 1694 & 0.20\\
    \rowcolor{blue!20} \cellcolor{white}& CG-\thres & 0.21 & 0.30 & 0.25 & 0.35 & \textbf{0.45} & 0.39 & 0.03 & 21 & 2.2 \\
    \rowcolor{blue!20} \cellcolor{white}& Khronos-\thres & 0.41  & 0.15  & 0.22 & \underline{0.72}  & 0.21 & 0.32 & \underline{0.11} & 162 & 0.26\\
    \rowcolor{blue!20} \cellcolor{white}& Clio-\batch & \textbf{0.52} & \textbf{0.34} & \textbf{0.41} & \underline{0.72} & \textbf{0.45} & \textbf{0.55} & \underline{0.11} & 90 & 0.23$^*$\\
    \rowcolor{blue!20} \cellcolor{white}& Clio-online & 0.35 & \underline{0.31} & \underline{0.33} & 0.52 & \underline{0.42} & \underline{0.46} & 0.07 & 99 & 0.26\\
    \midrule
    \multirow{-9}{*}{\rotatebox{90}{Apartment}}
    \end{tabular}}
    } %
    \vspace{-4mm}
    \caption{Results of locating objects of interest via open-set task \reda{queries for three datasets using 
        CLIP ViT-L/14}.
        The Office, Apartment, and Cubicle datasets have \reda{33}, 28, and 18 objects of interest respectively.
        \crossout{Results generated with 3090 GPU and Intel i9-12900K.}  %
        Shaded methods are informed by the list of tasks.
        First and second-best results are bolded and underlined, respectively.
        $^*$Total time for \name-batch normalized by number of images; 
    clustering step for batch run once on entire graph takes approximately 30 seconds and thus not suitable for online use. \label{table:three_d_results}}
\end{table}
 
\myParagraph{Results}
\reda{Firstly, we observe that task-informed approaches (shaded blue rows in~\cref{table:three_d_results}) 
lead to improved open-set precision and retain a much smaller amount of objects (``Objs'' column); motivating 
 our claim that metric-semantic mapping needs to be task-driven. In particular, in some cases \name retains 
an order of magnitude less objects compared to task-agnostic baselines (\cf with the number of objects in \name-Prim, 
which is essentially \name without the Information Bottleneck task-driven clustering).
We observe task-aware baselines, Khronos-\thres and \cg-\thres, have strictly worse open-set recall 
compared to their task-agnostic versions since both use awareness of the tasks to  
filter out irrelevant objects (improving open-set precision) 
but are unable to consider the tasks when forming objects (for example determining if a stack of notebooks is one object or multiple). 
This motivates our task-aware clustering approach as we observe 
that \name generally outperforms baselines across datasets and all metrics, with \name-batch and \name-online 
 ranking first or second in all but 2 cases, namely, the IOU and strict open-set recall metric in the Office dataset.}
 Many of the objects in the Office dataset (\eg staplers, bike helmet) are typically detected as isolated primitives, hence 
 we see that the knowledge of the task has a lesser impact on this dataset, while still improving performance across all other metrics.
 Third, we observe that \name is able to run in a fraction of a second and is around 6 times faster than \cg; 
 Khronos and \name-Prim also run in real-time, but have sub-par performance in terms of other metrics.
 Finally, \name-batch and \name-online have similar performance in most cases. Their performance difference is due to the fact that \name-online is executed in real-time and might drop frames as required to keep 
 up with the\crossout{camera} image stream. This difference sometimes helps and sometimes hinders the performance metrics.

As an example of \name's ability to 
use task information to form adequate
scene representations, 
\cref{fig:ib_tasks_visual}  shows a subset of the detected objects from \name for two different tasks sets.
For a task involving getting all condiment packets, \name represents a group of 
different type condiment packets collectively as one object,
while for an alternative set of tasks requiring specific types of condiments, \name represents the pile as multiple objects 
distinguished by sauce type, yielding a more flexible and useful scene representation. %
Qualitative results for the large-scale five-floor building dataset are included in the video attachment.

\subsection{\mbox{Closed-Set Object Evaluation}}
\label{sec:exp-closed-set-objects}

While \name is designed for open-set detection, we include results on 
the closed-set Replica~\cite{Straub19arxiv-replica} dataset 
using the evaluation method performed by~\cite{Jatavallabhula23rss-ConceptFusion, Gu24icra-conceptgraphs} 
to show that our task-aware mapping formulation does not degrade performance on closed-set mapping tasks. 
Here, our list of tasks is the set of object labels present in each Replica scene where each label is changed to be ``an image of \{class\}'' 
following~\cite{Gu24icra-conceptgraphs}. For both \name and~\cite{Gu24icra-conceptgraphs}, after creating the scene graph,
we assign the label with the highest cosine similarity to each of the detected objects.
To improve the reliability of CLIP given the low texture regions of the Replica dataset, we include global context CLIP 
vectors by incorporating dense CLIP features from~\cite{Shen23corl-F3RM} for \name.
We report accuracy as the class-mean recall (mAcc) and 
the frequency-weighted mean intersection-over-union (f-mIOU).
 \cref{table:replica} shows that \name achieves comparable performance to the leading \reda{methods on mAcc},
indicating that our task-aware clustering does not degrade performance on closed-set tasks. 
\reda{OpenMask3D~\cite{Takmaz23neurips-openmask3d} utilizes a 3D segmentation network which  
gives it superior performance in terms of f-mIOU but requires access to a full 3D reconstruction of the scene, limiting real-time application.}

\begin{table}[t]\scriptsize
    \vspace{1.5mm}
    \setlength{\tabcolsep}{17pt}
    \begin{tabular}{lcc}
        \toprule
        Method & mAcc & F-mIOU \\ 
        \midrule
        \addlinespace
        MaskCLIP \reda{~\cite{Ding23icml-maskclip}} & 4.53 & 0.94 \\ 
        \reda{Mask2former}~\reda{\cite{Cheng21cvpr-mask2former}} + Global CLIP feat  & 10.42 & 13.11 \\ 
        ConceptFusion \reda{~\cite{Jatavallabhula23rss-ConceptFusion}}  & 24.16 & 31.31 \\ 
        ConceptFusion \reda{~\cite{Jatavallabhula23rss-ConceptFusion}} + SAM & 31.53 & \reda{\underline{38.70}} \\ 
        \cg \reda{~\cite{Gu24icra-conceptgraphs}}  & \textbf{40.63} & 35.95 \\ 
        \cg-Detector \reda{~\cite{Gu24icra-conceptgraphs}} & 38.72 & 35.82 \\ 
        \reda{OpenMask3D~\cite{Takmaz23neurips-openmask3d}} & \reda{\underline{39.54}} & \reda{\textbf{49.26}} \\ 
        \name-\batch & 37.95 & 36.98 \\ 
        \bottomrule
    \end{tabular}
    \caption{Closed-set semantic segmentation experiments on 8 scenes from the Replica~\cite{Straub19arxiv-replica} dataset. 
    {Baseline results reported from~\cite{Gu24icra-conceptgraphs}.}  \label{table:replica} \vspace{0mm}}
\end{table}

\subsection{\mbox{Open Vocabulary Places Clustering}}
\label{sec:places_exp}

\reda{As} manually labeling \reda{open-set} 3D regions is a highly subjective task,
we evaluate the performance of Clio's regions via a proxy closed-set task, where \name{} is provided the set of possible room labels for the scenes as tasks.
We label rooms in three datasets: Office, Apartment, and Building.
We do not analyze the Cubicle or Replica~\cite{Straub19arxiv-replica} as they only consists of a single room.
We set $\alpha = 0$ to disable assignment to the null task as every place is relevant to at least one room label. \crossout{and we hold all parameters constant across scenes.}

\begin{table}[H]
    \scriptsize
    \centering
    \begin{tabularx}{\columnwidth}{lXccc}
        \toprule
        Dataset & Method & Precision$\uparrow$ & Recall$\uparrow$ & F1$\uparrow$ \\
        \midrule
        \multirow{4}{*}{Apartment} &            Hydra &    \underline{0.93 $\pm$ 0.01} &       \textbf{0.87 $\pm$ 0.01} &       \textbf{0.90 $\pm$ 0.00} \\
                                   &   Clio (closest) &                0.87 $\pm$ 0.06 &    \underline{0.78 $\pm$ 0.02} &    \underline{0.82 $\pm$ 0.01} \\
                                   &   Clio (average) &       \textbf{0.98 $\pm$ 0.02} &                0.54 $\pm$ 0.00 &                0.69 $\pm$ 0.00 \\
        \midrule
        \multirow{4}{*}{Office}    &            Hydra &                0.61 $\pm$ 0.03 &       \textbf{0.84 $\pm$ 0.03} &                0.70 $\pm$ 0.01 \\
                                   &   Clio (closest) &    \underline{0.67 $\pm$ 0.03} &                0.79 $\pm$ 0.01 &    \underline{0.72 $\pm$ 0.01} \\
                                   &   Clio (average) &       \textbf{0.73 $\pm$ 0.01} &    \underline{0.80 $\pm$ 0.00} &       \textbf{0.76 $\pm$ 0.01} \\
        \midrule
        \multirow{4}{*}{Building}  &            Hydra &       \textbf{0.87 $\pm$ 0.01} &                0.71 $\pm$ 0.02 &    \underline{0.78 $\pm$ 0.01} \\
                                   &   Clio (closest) &                0.72 $\pm$ 0.04 &                0.82 $\pm$ 0.01 &                0.77 $\pm$ 0.02 \\
                                   &   Clio (average) &    \underline{0.79 $\pm$ 0.02} &       \textbf{0.84 $\pm$ 0.01} &       \textbf{0.81 $\pm$ 0.01} \\
        \bottomrule
    \end{tabularx}
    \caption{Comparison of geometric room segmentation accuracy.}\label{table:places}
\end{table}
 
We use the precision and recall metrics presented \reda{in}~\cite{Hughes24ijrr-hydraFoundations} \reda{to assess the geometric accuracy of the predicted rooms of}
our proposed CLIP embedding vector association strategy, \emph{\name{} (average)}.
\reda{We compare with an alternative strategy, \emph{\name{} (closest)}, which uses the embedding vector taken from the closest image that the place node is visible from,
and the} purely geometric room segmentation approach from Hydra~\cite{Hughes24ijrr-hydraFoundations}%
\crossout{as a point of comparison for the closed set performance}.
Results from this comparison are presented in~\cref{table:places}, which also includes the F1 score as a summary statistic.
The results in~\cref{table:places} are averaged over 5 trials, and standard deviation of all metrics is reported.
We note that our chosen association strategy outperforms both the purely geometric approach of Hydra~\cite{Hughes24ijrr-hydraFoundations} and the more naive \emph{\name{} (closest)} for the Office and Building scene, but performs relatively poorly in terms of F1 score in the Apartment. %
This is due to the nature of the scenes; the Office and the Building scene contain labeled open floor-plan rooms that require semantic knowledge to be detected (\eg{} a kitchenette in the Office scene or stairwells in the Building scene).
The Apartment primarily contains geometrically distinct rooms, which are straightforward to segment with the geometric approach in~\cite{Hughes24ijrr-hydraFoundations}, and are instead over-segmented by \name, as evident from the  high precision but low recall of our method.
On the other hand, semantically similar regions that are connected, as present in the Office, lead to under-segmentation and lower recall compared to Hydra~\cite{Hughes24ijrr-hydraFoundations}.

\begin{figure}[tb]
    \vspace{1.5mm}
    \centering
    \includegraphics[width=0.49\columnwidth, trim={9cm 1cm 12cm 1cm}, clip, angle=0, origin=c]{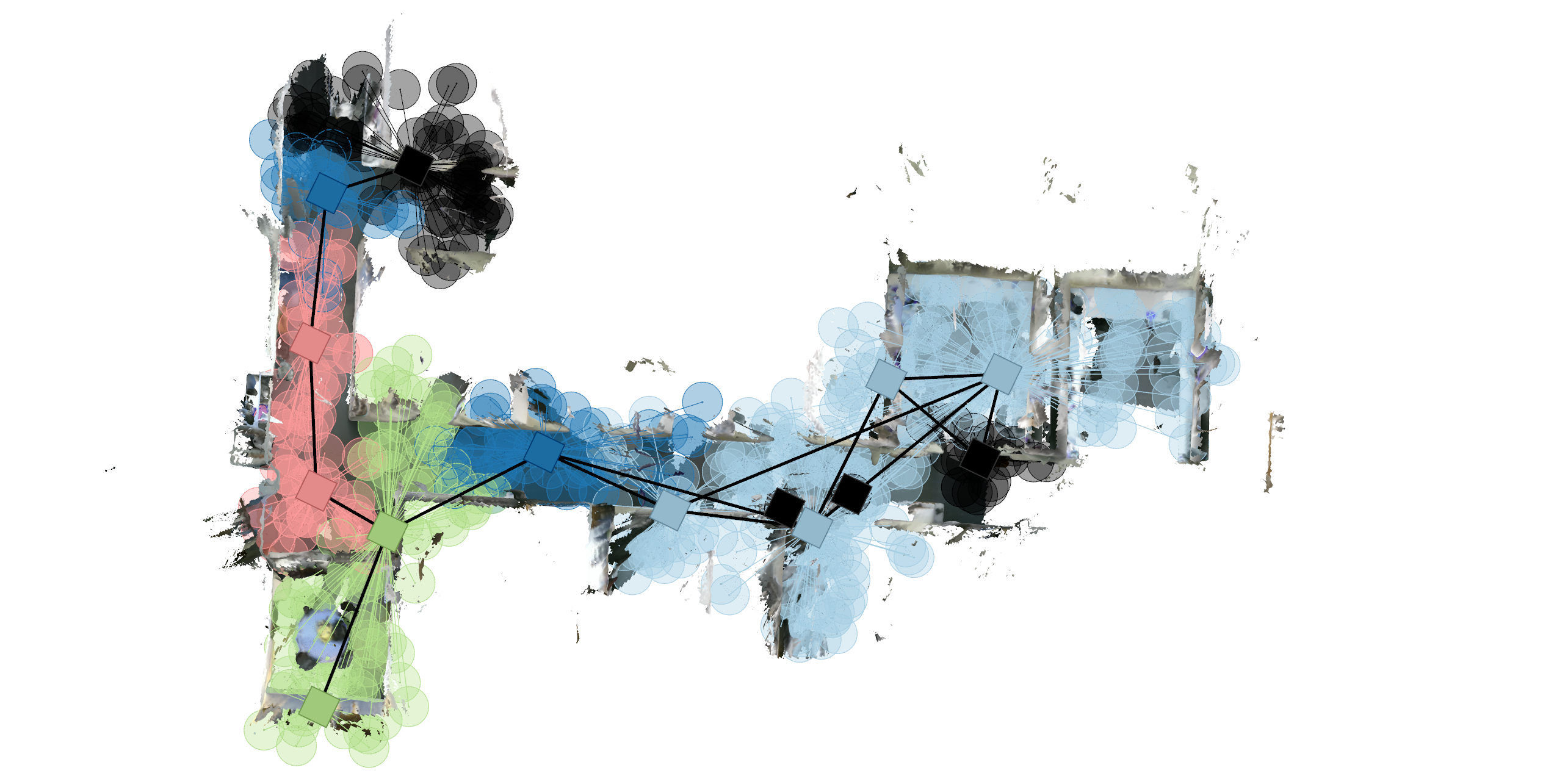}
    \includegraphics[width=0.49\columnwidth, trim={9cm 1cm 12cm 1cm}, clip, angle=0, origin=c]{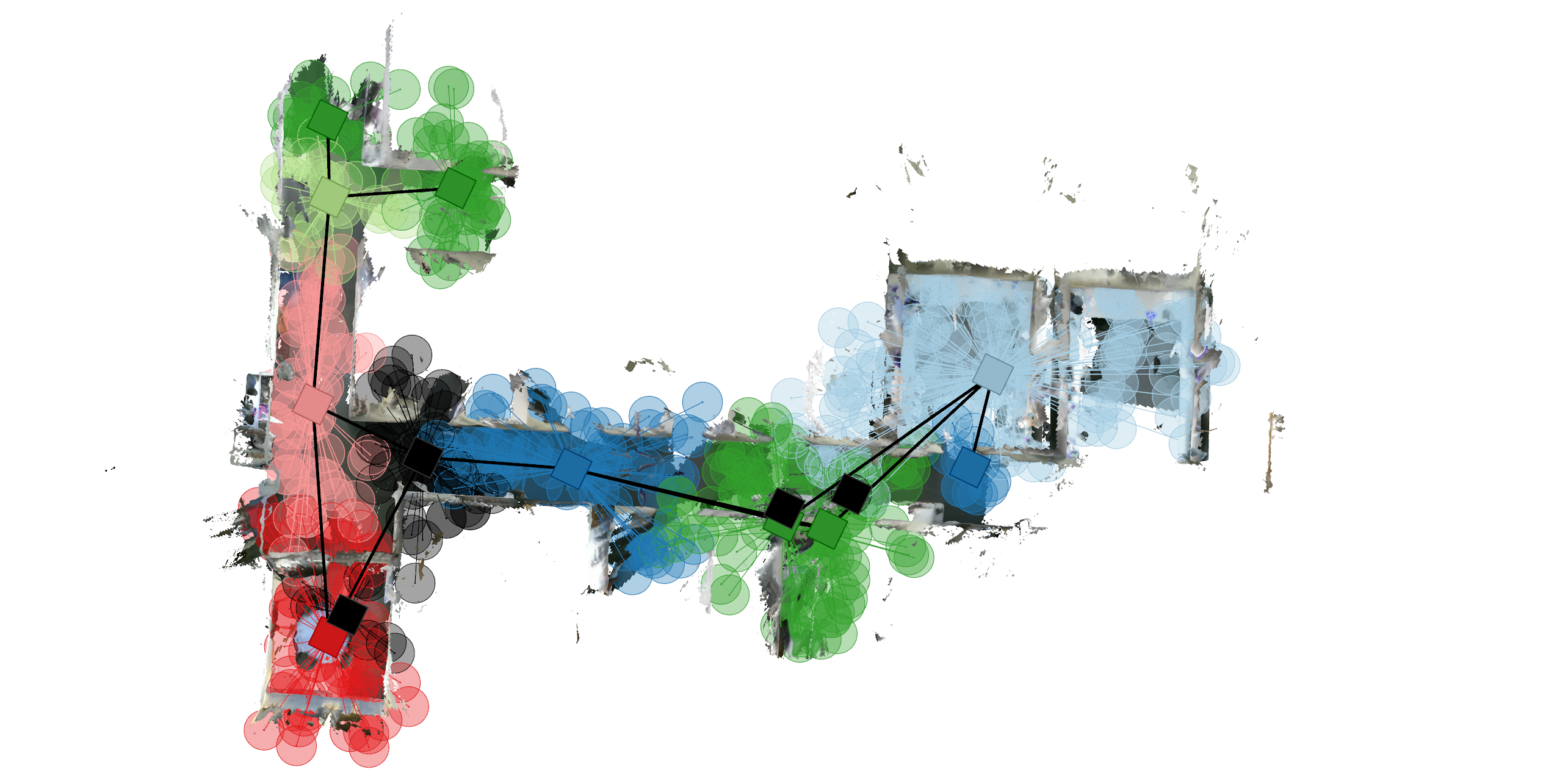} \\
    \includegraphics[width=0.41\columnwidth]{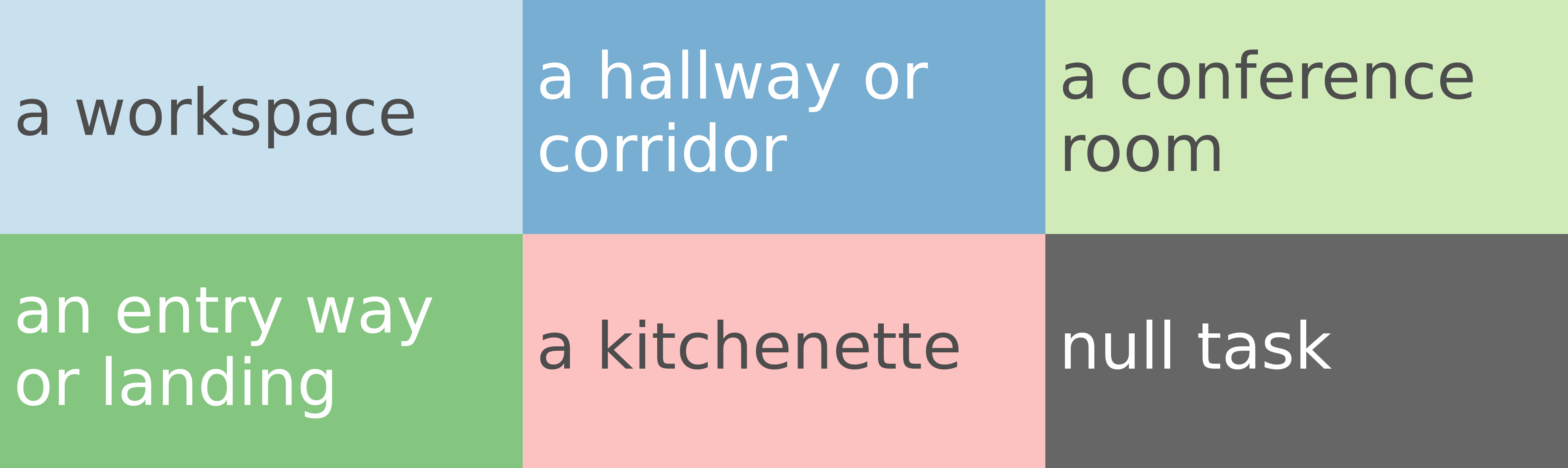}
    \includegraphics[width=0.57\columnwidth]{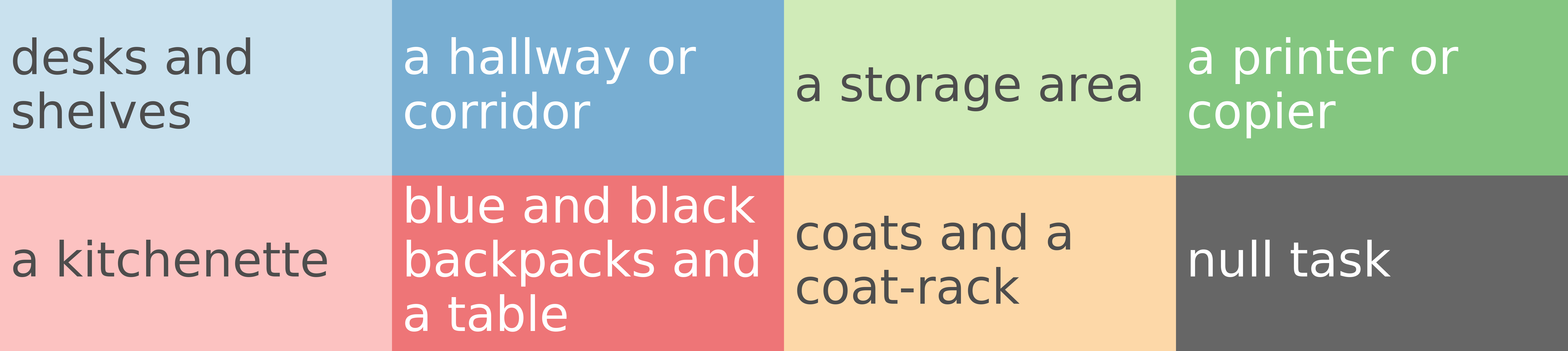}
    \caption{%
        Qualitative examples of places clustering.
        The first figure shows regions that result from clustering by task prompts resembling room category labels.
        The second figure shows regions that result from clustering by task prompts that are a mix of potential rooms and objects.
    }\label{fig:qualitative_places}
\end{figure}
 
\cref{fig:qualitative_places} qualitatively demonstrates \name{}'s capability  to produce task-relevant regions on the Office scene. %
We compare two different granularities of tasks; the first is similar to the provided room labels in the closed-set proxy evaluation while the second is\crossout{much} more granular and object-driven.
The resulting regions reflect this difference in granularity despite being produced by \name using the same set of parameters.
More visualizations supporting the meaningfulness of \name's region clustering are provided in \cref{sec:places_ap_viz}.
\subsection{\mbox{Online Evaluation on Spot}}
\label{sec:spot}

To demonstrate the real-time use of \name for robotics, we conduct mobile manipulation experiments using a Boston Dynamics Spot quadruped robot equipped with an arm and gripper.
During the experiments, the robot constructs a map with \name in real-time while exploring a scene, and then is tasked to navigate to and pick up objects matching a provided natural language prompt (\eg{} \cref{fig:spot}).
We then compute the shortest path through the place nodes to the target object via Dijkstra's algorithm.\crossout{We smooth and command Spot to follow the resulting trajectory to navigate to the object by commanding Spot to sequentially navigate to sampled waypoints.}
After reaching the target object, we select the pixel centroid from the current input semantic segments with the highest cosine similarity to the prompt embedding \crossout{as a grasp candidate}as input to the Spot API grasp command.
We use the onboard front-left and front-right RGB-D cameras and odometry from Spot as inputs to \name.
We run \name on a laptop capable of being mounted on the robot that is equipped with an Intel i9-13950HX CPU with $24$ cores, $64$GB of RAM, and an NVIDIA GeForce RTX $4090$ Laptop GPU. %

\begin{figure}[tb]
    \vspace{1.5mm}
    \centering
    \includegraphics[width=\columnwidth, trim={1.0cm 0.20cm 1.0cm 0.65cm}, clip]{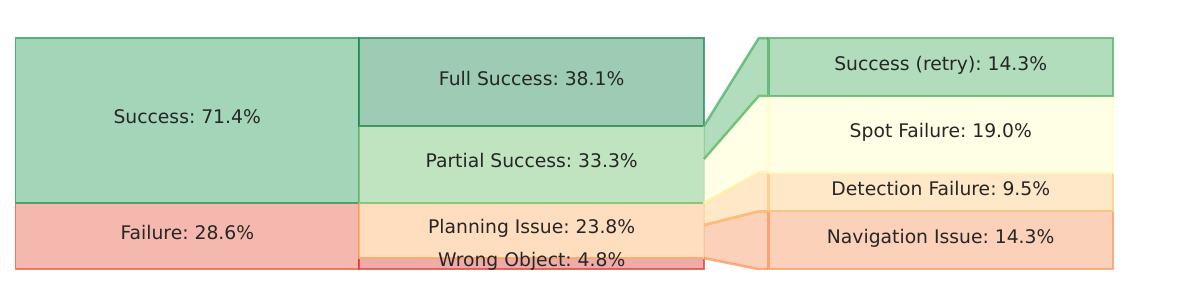}
    \caption{%
        Breakdown of grasp results for the 21 object grasp attempts performed by Spot.
        ``Wrong object'' refers to the wrong \name object being selected,
        ``Detection failure'' refers to the selected image coordinates for grasping not corresponding to the target object, ``Navigation issue'' refers to the trajectory resulting in a pose where the object was not visible, ``Spot Failure'' refers to the Spot API %
        failing to pick up a correctly identified grasp, and ``Success (retry)'' refers to the Spot API grasp command failing to pick up the object on the first attempt but succeeding after repeated attempts.
    }\label{fig:spot_trials}
\end{figure}

\crossout{For these experiments, each command is embedded using CLIP and used to select the object in the scene graph with the highest cosine similarity to the prompt.}
We perform 7 trials of a mobile manipulation experiment.
\crossout{where we use \name and our planner described above to grasp a variety of objects.}\footnote{%
We consider 7 different objects for grasping: a rope dog toy, a snorkel, a stuffed animal, a backpack, a measuring tape, a water bottle, and two different colored plastic cones.
Trials are performed with the laptop off-board and connected to Spot via WiFi due to logistical challenges (\eg{} battery life) inherent in repeated manipulation trials, while the video attachment shows an uninterrupted experiment with onboard computation.
}
Each trial consists of a mapping phase \reda{and a planning phase. In the mapping phase} we teleoperate Spot to observe all the objects in the scene (consisting of two room-like areas joined by a hallway).
After the mapping phase, we move Spot to a starting location \reda{for the planning phase where we} command grasps of 3 random target objects for a total of 21 unique grasp attempts.
\name runs the entire time during \reda{each} trial, and no post-processing of the 3D scene graph is performed.
We present a breakdown of the 21 trials in~\cref{fig:spot_trials}.
Overall, we achieve a 57\% success rate for the grasps and a 71\% success rate if we disregard the cases where Spot failed to actually grasp a correctly identified object.
Notably, \name was only unable to select the correct target object in the scene graph once (\ie{} the ``Wrong Object'' failure category).
The video attachment also demonstrates a pick-and-place experiment with a sequence of 4 pick-and-place actions over a larger area where Spot is operated with the laptop onboard.
These experiments together emphasize the suitability of \name for use \reda{on board} real robotic platforms.

\section{Limitations}
\label{sec:limits}
Despite the encouraging experimental results, our approach has multiple limitations.
First, while our method is zero-shot and is not bound to any particular foundation model, it does 
inherit some limitations from the foundation models used in implementation such as strong vulnerability to prompt tuning. 
For instance, in \cref{sec:open_clip_objects}, we discuss how performance is affected by different CLIP models.
Second, we currently average CLIP vectors when merging two primitives, but it would be interesting to 
consider more grounded ways to combine\crossout{their} semantic descriptions.
Third, \name can over-cluster if two primitives individually have similar cosine similarity to the same task 
but\crossout{somehow} the task requires distinguishing them as separate objects (\eg we might want to distinguish a fork from a knife when setting \reda{a} table, even though they might have similar relevance to the task).
Finally, we currently 
consider relatively simple, single-step tasks.
However, it would be desirable to extend the proposed framework 
to work with a set of high-level, complex tasks, \reda{including tasks that require substantial understanding of object parts.}

\section{Conclusion}
\label{sec:conclusion}

We have presented a task-driven formulation for 3D metric-semantic mapping, where a robot is provided 
with a list of natural language tasks and has to create a map whose granularity and structure is sufficient to support those tasks. 
We have shown that this problem can be expressed in terms of the classical Information Bottleneck and have 
developed an incremental version of the Agglomerative Information Bottleneck algorithm as a solution strategy. 
We have integrated the resulting algorithm in a real-time system, \name, that 
constructs a 3D scene graph ---including task-relevant objects and regions--- as the robot explores the environment.
We have also demonstrated \name's relevance for robotics, by showing it can be executed in real-time onboard  a Spot robot and  support pick-and-place mobile manipulation tasks. %

\section*{Acknowledgement}
\reda{We would like to acknowledge Bryan Zhao for the help with prototyping a trajectory planner on 3D scene graphs.}
\crossout{The authors would like to acknowledge Bryan Zhao for his help in prototyping an earlier version of code to perform trajectory planning on 3D scene graphs for a prior project.}

{\scriptsize
\bibliographystyle{IEEEtran}

}

\appendix
\label{sec:appendix}

\subsection{Agglomerative Information Bottleneck}
\label{sec:aib}

\cref{alg:aig} provides the pseudocode for the Agglomerative Information Bottleneck~\cite{Slonim99nips-AgglomerativeIB} 
discussed in Section~\ref{sec:obj_clustering}. The goal of \cref{alg:aig} is to find an optimal hard clustering 
assignment $p(\tilde{x}|x)$ that compresses an initial signal $X$ into a compressed signal $\tilde{X}$ while preserving 
relevant information about a relevancy variable $Y$ (which in our case is a set of tasks).
The algorithm runs until a set 
threshold $\bar{\delta}$ is reached which is used to regulate the amount of compression with respect to preserving information about $Y$. 

\begin{algorithm}
    \caption{Agglomerative Information Bottleneck}
    \label{alg:aig}
    \begin{algorithmic}[1]
    \renewcommand{\algorithmicrequire}{\textbf{Input:}}
    \renewcommand{\algorithmicensure}{\textbf{Output:}}
    \REQUIRE$\bar{\delta}$, initial primitives $\{x_1, \ldots x_N\} = X$, task-list $Y$
    \ENSURE  $p(\tilde{x}|x)$: hard assignment of primitives to clusters
    \\ {\color{gray} \% Initialization:}
    \STATE set $p(y|x)$ using $eq.~\eqref{eq:pyx}$
     \STATE $\tilde{x}_i = x_i \ \forall x_i \in X$
     \STATE {$p(\tilde{x}_i) = p(x_i) = 1/N$} {\color{gray} \% uniform distribution}
     \STATE $p(y|\tilde{x}) = p(y|x_i), \forall y \in Y$
     \STATE Compute $d_{ij}$ using eq.~\eqref{eq:distortion} for all $i = 1,\ldots,|X|$ and $j = 1,\ldots,|Y|$
    \\ {\color{gray} \% Main loop:}
     \WHILE {$\delta < \bar{\delta}$}
     \STATE $d_{ab} = \min_{ij}(d_{ij})$ %
     \STATE $p(\tilde{x}) = p(x_a) + p(x_b)$
     \STATE $p(y|\tilde{x}) = \frac{p(x_a, y) + p(x_b, y)}{p(\tilde{x})}\ \forall y \in Y$
     \STATE $p(\tilde{x}|x)  = 1\ if\ x \in \tilde{x}_a \cup \tilde{x}_b,\ 0 \ \text{ otherwise } \ \forall x \in X$
     \STATE compute $\delta$ from eq.~\eqref{eq:delta} for batch or eq.~\eqref{eq:delta_reweighted} for online
     \ENDWHILE
    \RETURN $p(\tilde{x}|x)$
    \end{algorithmic} 
    \end{algorithm}

\subsection{Incremental Agglomerative IB}
\label{sec:iaib_proof}
As mentioned in~\cref{sec:obj_clustering}, we form an incremental version 
of the Agglomerative \ib to run \name online. 
For this, we run Agglomerative \ib on each individual connected component $c$ using a re-weighted definition of $\delta(k)$. 
Assuming that $p(x)$ is a uniform distribution we can write the incremental equivalent of $\delta(k)$ as: 
\begin{equation}\label{eq:delta_reweighted}
	\delta_c(k) = \frac{|X_c|}{|X|}\frac{I((\tilde{X_c})_k;Y) - I((\tilde{X_c})_{k-1} ; Y)}{I(X;Y)}
\end{equation}
where $X_c$ are the primitives in component $c$. This gives the exact same result as Agglomerative \ib on the full graph  
which lets us implement the stopping 
condition of \cref{alg:aig} across each connected component
Therefore, we can solve Agglomerative \ib in an incremental manner 
by only performing Agglomerative \ib on the subset of connected components of the graph that are affected by new measurements 
using \cref{alg:iaig}. Here, when \name receives new primitives $X_{new}$, we add the primitives to their respective sub-graphs 
and for each of the sub-graphs that received new primitives we run Agglomerative \ib until the stopping condition from 
eq.~\eqref{eq:delta_reweighted} is met, repeating as new primitives are received.

\begin{algorithm}
    \caption{Incremental Agglomerative Information Bottleneck}
    \label{alg:iaig}
    \begin{algorithmic}[1]
    \renewcommand{\algorithmicrequire}{\textbf{Input:}}
    \renewcommand{\algorithmicensure}{\textbf{Output:}}
    \REQUIRE $c \in \calC$ \COMMENT{set of connected sub-graphs}\\
    \ \ \ \ $X_{new}$ \COMMENT{newly received primitives}
    \ENSURE  $p(\tilde{x}|x)$: hard assignment of primitives to clusters
    \STATE $\calC \leftarrow X_{new}$ \COMMENT {update corresponding sub-graphs with new primitives}
    \FOR{each $c$ in $\calC$}
    \IF{$c$ updated}
    \STATE update $p(\tilde{x}|x), x \in X_c,\ \tilde{x} \in \tilde{X}_c $ with \cref{alg:aig} using stop condition from eq.~\eqref{eq:delta_reweighted}
    \ENDIF
    \ENDFOR
    \RETURN $p(\tilde{x}|x)$
    \end{algorithmic} 
    \end{algorithm}

Here we provide the proof to the expression in eq.~\eqref{eq:delta_reweighted}.
Given a connected component $c$ we want to cluster $X_c$, the primitives within the component, into clusters $\tilde{X_c}$
independent of the rest of the graph.
Let us also define $o$ for the primitives not in $c$ such that $X_c \cup X_o = X$ and $X_c \cap X_o = \varnothing$.
Since $P(X)$ is uniformly distributed,
\begin{equation}
I(X_c;Y) = \frac{1}{|X_c|} \sum_{X_c} p(y|x) \log(\frac{p(y|x)}{p(y)})
\end{equation}

Let us define $\Delta$ such that
\begin{equation}
\Delta = \frac{1}{|X|} \sum_{X_o} p(y|x) \log(\frac{p(y|x)}{p(y)})
\end{equation}
this allows us to rewrite $I(X; Y)$ as follows:
\begin{equation}
\begin{aligned}
I(X;Y) = & \frac{1}{|X|} \sum_{X_c} p(y|x) \log(\frac{p(y|x)}{p(y)}) + \Delta \\
       = & \frac{|X_c|}{|X|} I(X_c;Y) + \Delta
\end{aligned}
\end{equation}
since we are only clustering in $c$,
\begin{equation}
I(\tilde{X}_k;Y) = \frac{|X_c|}{|X|} I((\tilde{X}_c)_k;Y) + \Delta
\end{equation}

Substituting in for $I(\tilde{X}_k;Y)$ and $I(\tilde{X}_{k-1};Y)$ in \eqref{eq:delta},
we obtain our re-weighted expression in \eqref{eq:delta_reweighted}.

\subsection{Office, Apartment, and Cubicle Datasets}
\label{sec:datasets}

\begin{figure}[h]
    \centering
    \begin{subfigure}[b]{0.47\textwidth}
        \includegraphics[width=\textwidth]{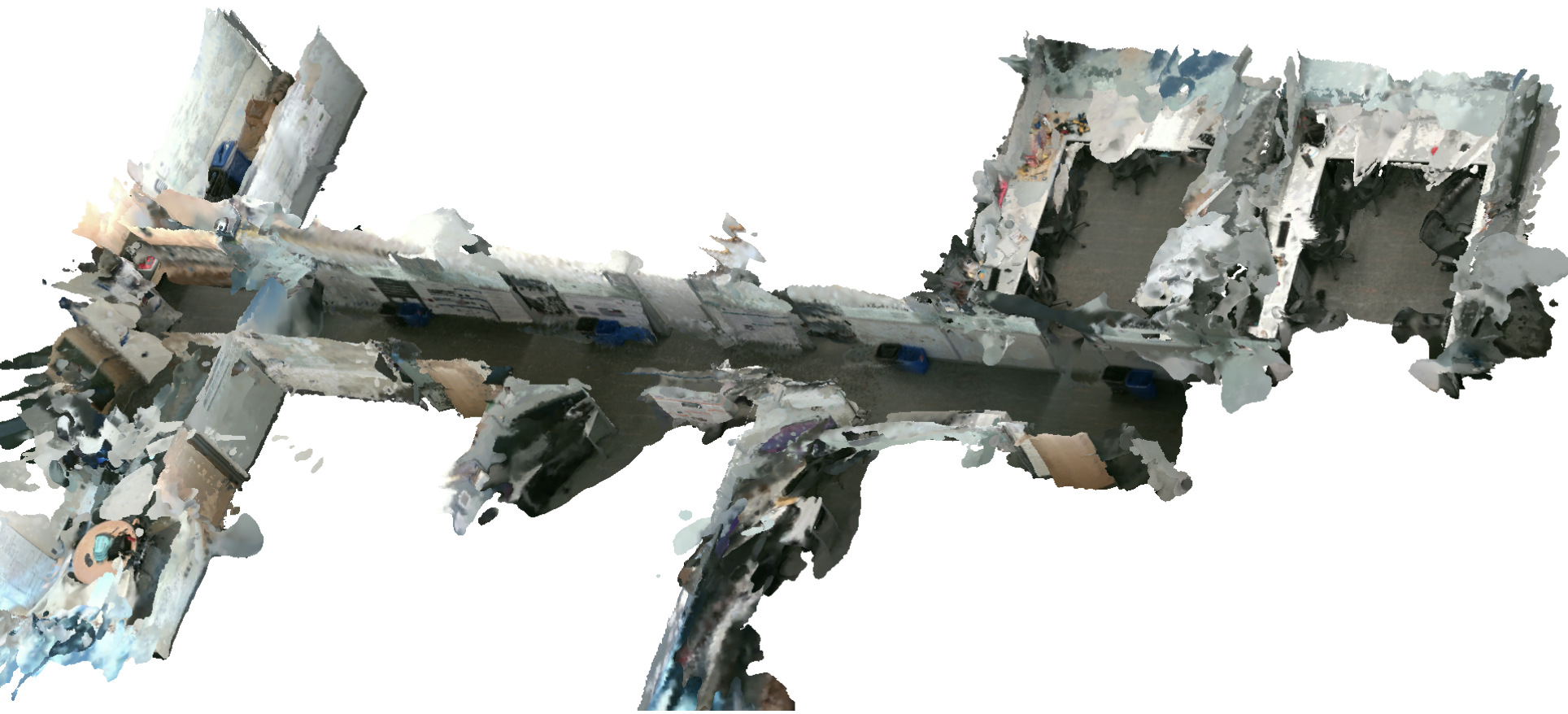}
        \caption{Office Scene}
        \label{fig:scene_o}
    \end{subfigure}
    \hspace{\fill} %
    \centering
    \begin{subfigure}[b]{0.47\textwidth}
        \includegraphics[width=\textwidth]{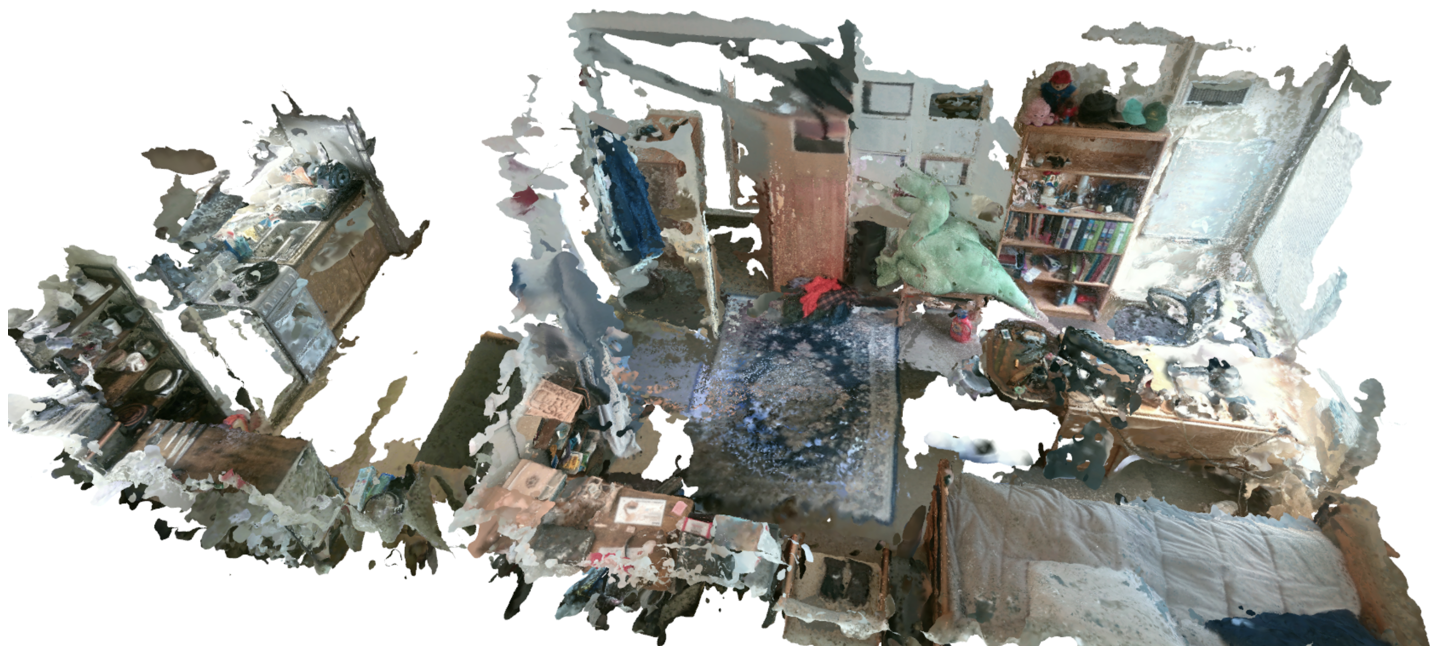}
        \caption{Apartment Scene}
        \label{fig:scene_a}
    \end{subfigure}
    \hspace{\fill} %
    \begin{subfigure}[b]{0.47\textwidth}
        \includegraphics[width=\textwidth]{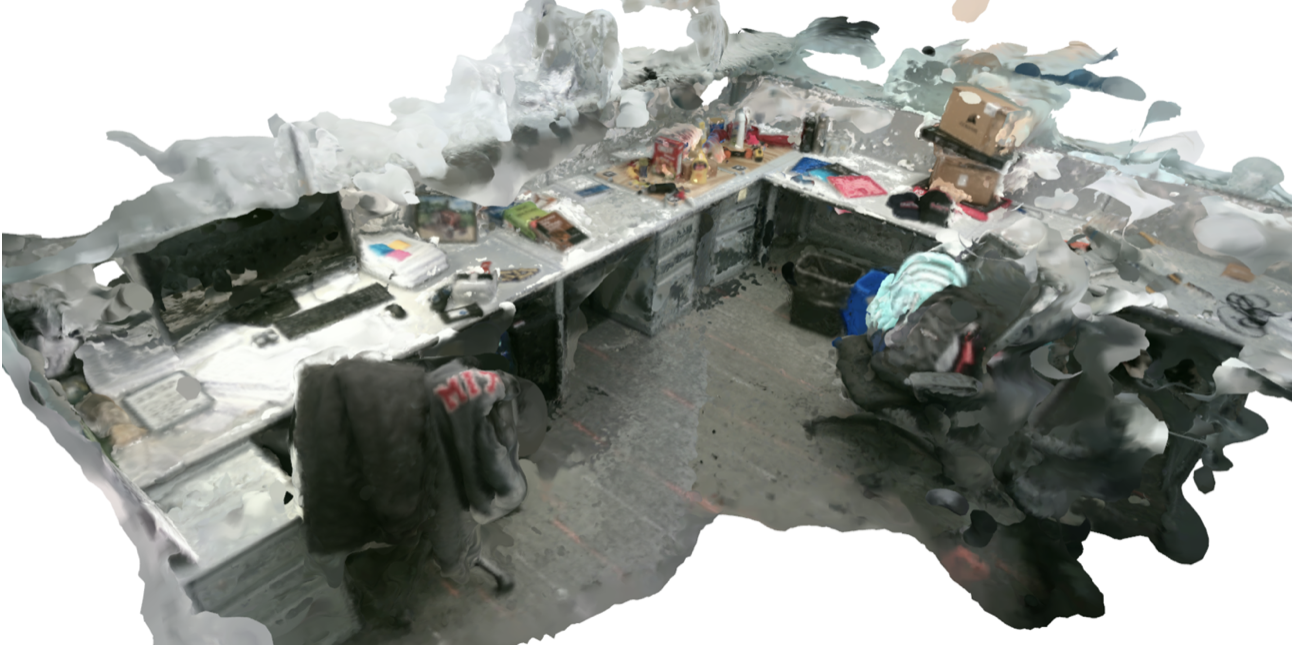}
        \caption{Cubicle Scene}
        \label{fig:scene_c}
    \end{subfigure}
    \caption{Custom open-vocabulary 3D datasets of an office floor, apartment, and cubicle. }
    \label{fig:scene_meshes}
\end{figure}

For each of the office, apartment, cubicle and building datasets, we collect RGB-D images with an Intel RealSense D455. 
A visualization of the scenes are shown in \cref{fig:scene_meshes}.

\begin{figure*}[h]
    \centering
    \begin{subfigure}[b]{\textwidth}
        \centering
        \includegraphics[width=\textwidth, trim={1cm 3cm 1cm 6cm}, clip]{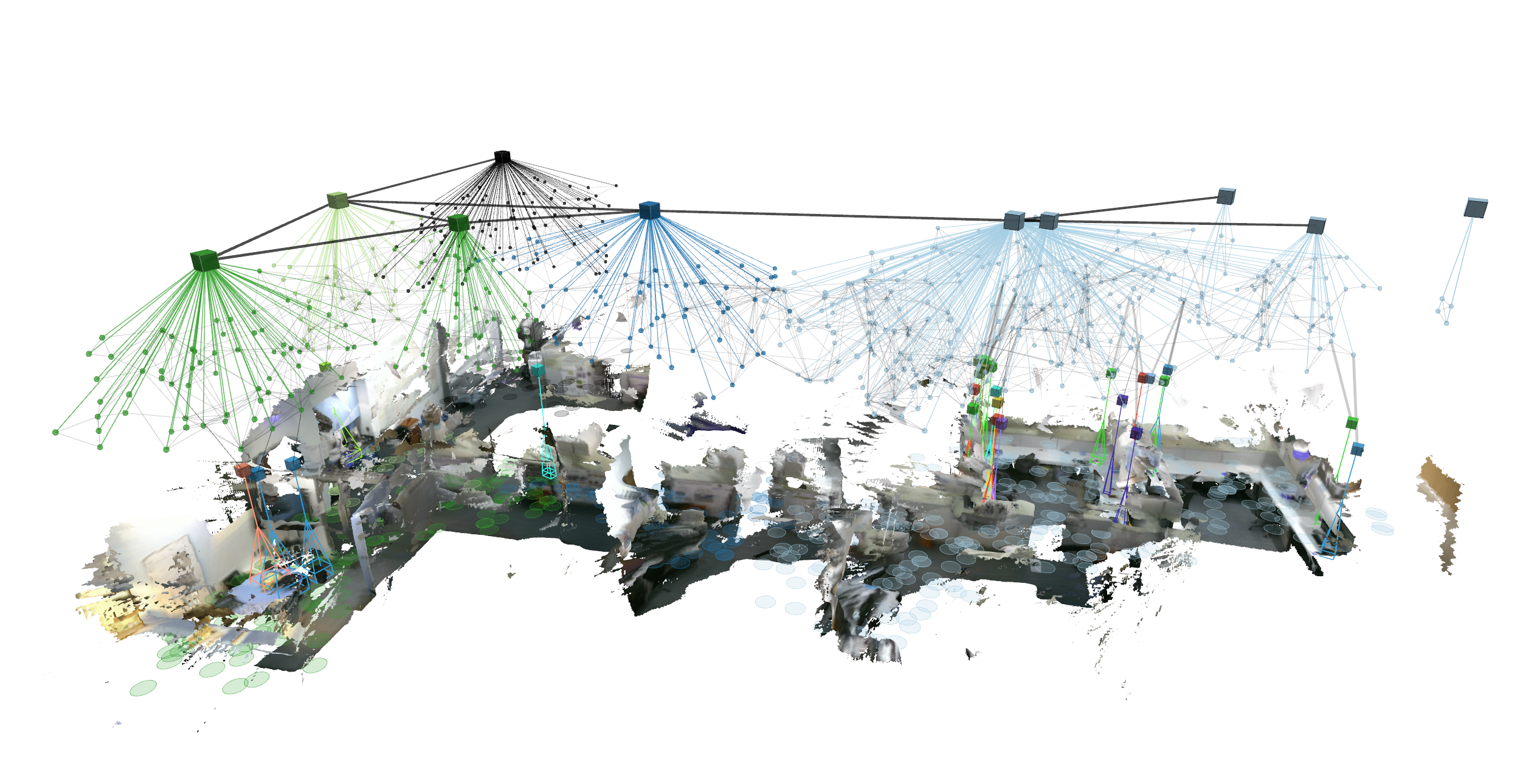} \\
        \includegraphics[width=0.5\textwidth]{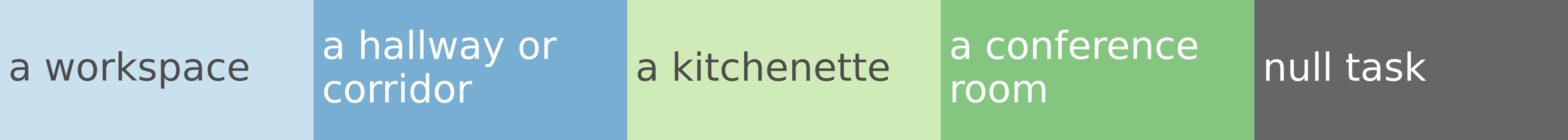}
        \caption{Office Scene}
    \end{subfigure}
    \begin{subfigure}[b]{0.48\textwidth}
        \centering
        \includegraphics[width=0.7\textwidth, trim={22cm 1cm 15cm 0cm}, clip]{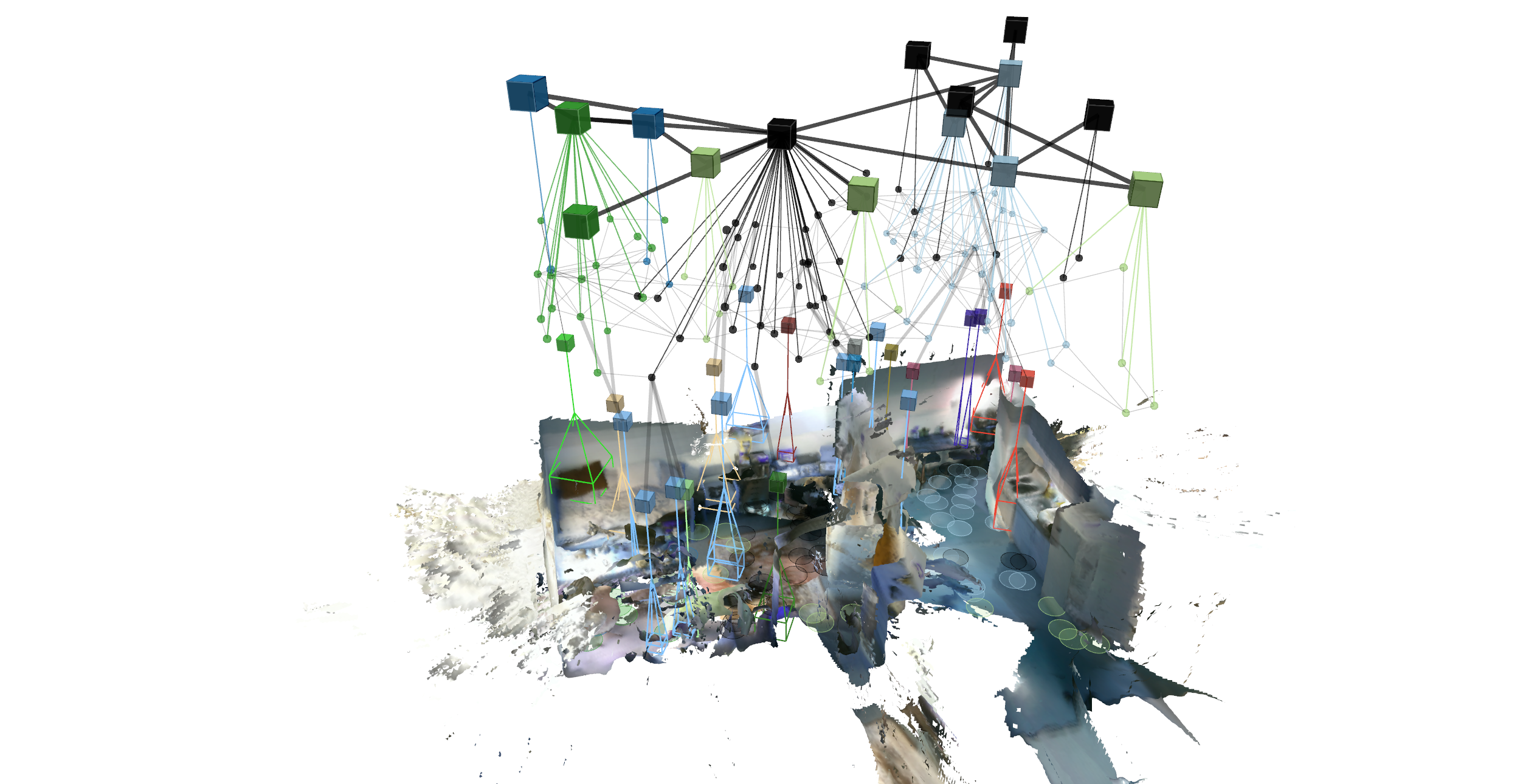} \\
        \includegraphics[width=\textwidth]{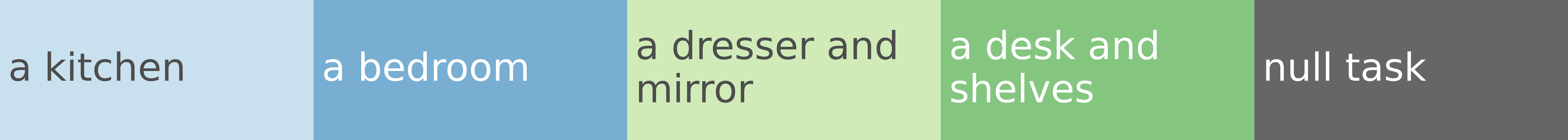}
        \caption{Apartment Scene}
    \end{subfigure}
    \hspace{\fill} %
    \begin{subfigure}[b]{0.48\textwidth}
        \centering
        \includegraphics[width=0.7\textwidth, trim={20cm 1cm 20cm 1cm}, clip]{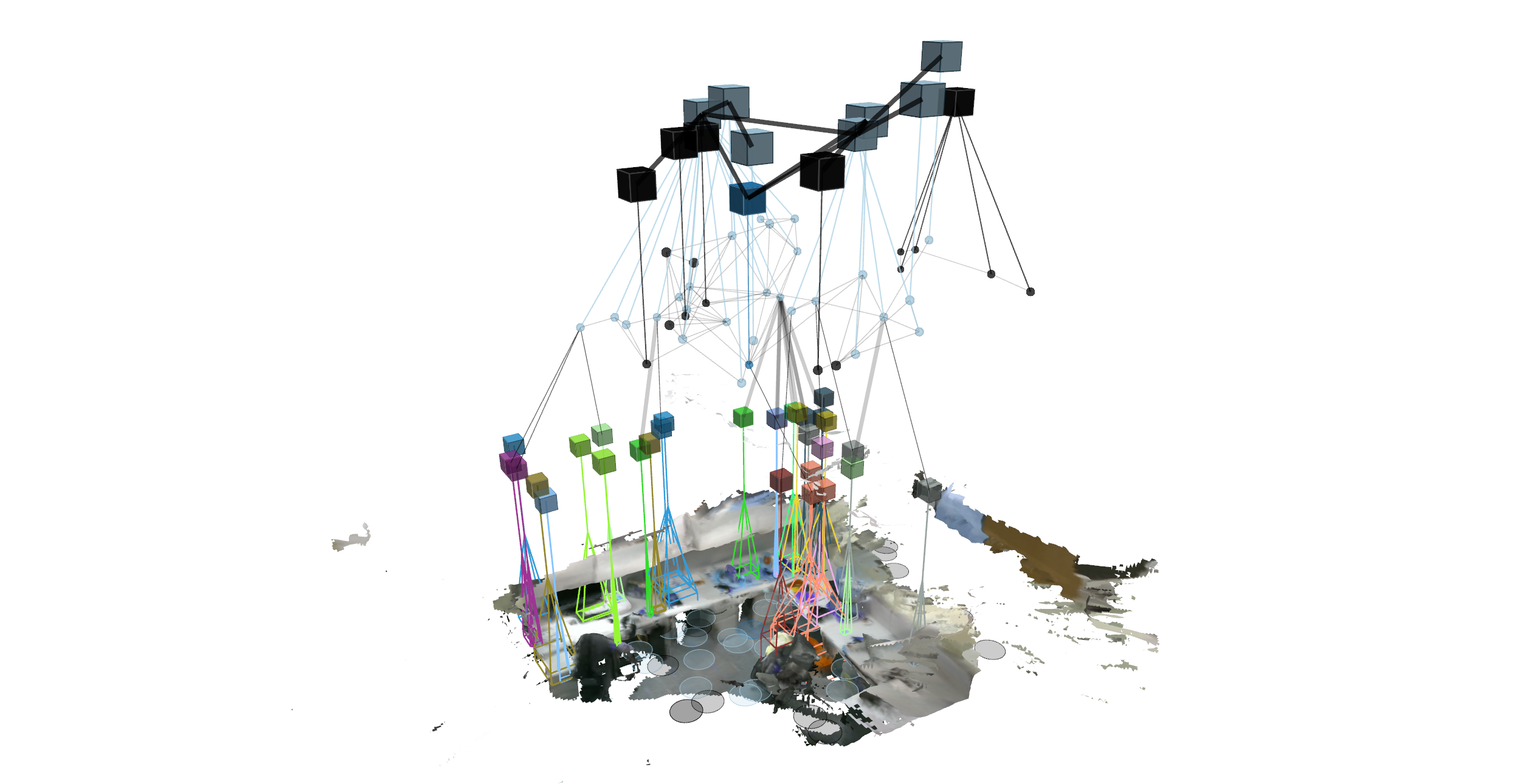} \\
        \includegraphics[width=0.6\textwidth]{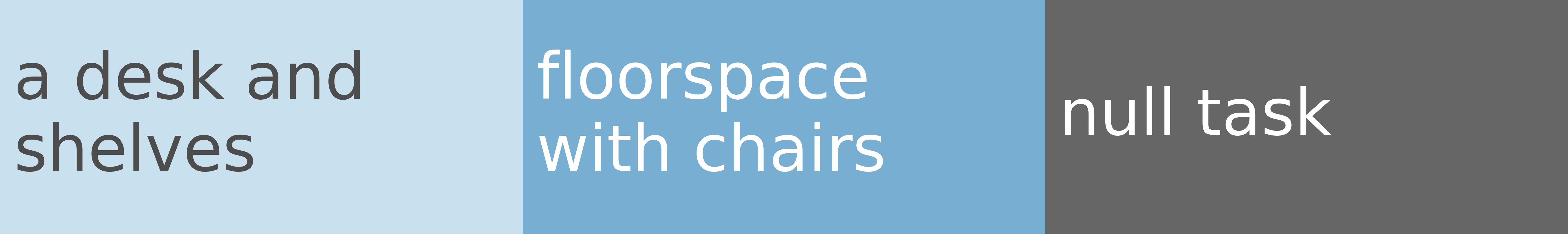}
        \caption{Cubicle Scene}
    \end{subfigure}
    \centering
    \caption{%
        Example 3D scene graphs for the self-collected Office, Apartment and Cubicle datasets.
        Scene graphs layers are drawn in the following order: objects (as cubes), places (as spheres) and regions (as cubes).
        The bounding box of each object is drawn below, and a footprint is drawn for each place primitive to highlight the 2D positions of the nodes.
        Places and regions are colored by their closest task as shown in the legend below each figure.
    }\label{fig:scene_dsgs}
\end{figure*}

A visualization of the resulting scene graphs are also shown in \cref{fig:scene_dsgs}.

\subsection{Office Scene Task List}
\label{sec:office_labels}
Here we provide a list of tasks used during mapping and querying of the office scene. The number of 
objects assigned to each task is included in parentheses. There are 33 distinct objects in total.

\begin{enumerate}
\item get a black Expo marker (2)
\item get a painting of a tractor (1)
\item move rack of magazines (1)
\item get my Signals and Systems textbook (1)
\item something to cut paper (3)
\item get black glasses (1)
\item get box of tissues (2)
\item get my gloves (1)
\item get orange knit hat to keep my head warm (1)
\item get rock with holes (1)
\item something to put on a hot dog (1)
\item get can of tuna (1)
\item grab black backpack (1)
\item grab teal backpack (1)
\item move the bin of clothes (1)
\item move the printer (3)
\item organize the pile of red dishes and plates (1)
\item get stapler (2)
\item get a yellow rubber duck (1)
\item organize the pile of hardware tools (1)
\item count solid core wood doors (3)
\item polish metal lever handle and sideplate (3)
\end{enumerate}

\subsection{Apartment Scene Task List}
\label{sec:apartment_labels}
Here we provide a list of tasks used during mapping and querying of the apartment scene. The number of 
objects assigned to each task is included in parentheses. There are 28 distinct objects in total. 

\begin{enumerate}
\item get can of WD-40 (1)
\item clean toaster (1)
\item find deck of cards (1)
\item find pile of hats (1)
\item find spice bottles (1)
\item get a kitchen knife (3)
\item get pocket knife (1)
\item get bike helmet (1)
\item get bottle of tide (1)
\item get cast iron skillet (1)
\item get hair dryer (1)
\item get hairbrush (1)
\item get notebooks binders (1)
\item get pizza cutting wheel (1)
\item get soy sauce (1)
\item get toolbox (1)
\item get violin case (1)
\item move pile of clothes (1)
\item move rack of dishes (1)
\item bring me a pillow (2)
\item get alarm clock (1)
\item get all chocolate snacks (1)
\item get chapstick (1)
\item get first aid kit (1)
\item move popcorn bags (1)
\end{enumerate}

\subsection{Cubicle Scene Task List}
\label{sec:cubicle_labels}
Here we provide a list of tasks used during mapping and querying of the cubicle scene. All tasks here have
one corresponding object. There are 18 objects in total.

\begin{enumerate}
\item get condiment packets
\item get drink cans 
\item get eyeglasses 
\item get glasses case 
\item get grey jacket 
\item get my silver water bottle 
\item get notebooks 
\item get mudstone rock
\item tool to cut paper 
\item get sticky notes 
\item get textbooks 
\item get waste bins 
\item move hats 
\item clean backpacks 
\item get red crockery 
\item get hardware drill 
\item get quartz rock 
\item get tape measure
\end{enumerate}

\subsection{Building Scene Task List}
\label{sec:builing33_labels}
Here we provide a list of tasks used during mapping and querying of the building scene. Note that 
some tasks have many occurrences of relevant items in the dataset.

\begin{enumerate}
    \item get Lysol
    \item get vacuum cleaner
    \item get fire extinguisher
    \item get yellow wet floor sign
    \item get clamps
    \item get epoxy and resin bottles
    \item get roles of tape
    \item locate screwdrivers
    \item move jet engine
    \item get earmuffs 
    \item move co2 tanks
    \item check office printer
    \item get books
    \item get basketball
    \item refill dish soap bottles
    \item get trashbins
    \item move pink foam
    \item stack blue foam
    \item check microwave
    \item clean sink
    \item get bottles of cleaner
    \item stuff with MIT on it
    \item get tape measure
    \item grab airplane wing
    \item clean stairs
\end{enumerate}

\subsection{Open Vocabulary Tasks on OpenCLIP model}
\label{sec:open_clip_objects}
Here we repeat the experiments from \cref{table:three_d_results} but this time use a different CLIP 
model (ViT-H-14 from OpenCLIP~\cite{Ilharco21github-openclip}). Due to the higher compute requirements for this model 
we do not run \name-online and instead only run \name-batch. We found that this model tends to produce higher cosine similarity scores 
between image primitives and tasks for both relevant and irrelevant pairings, and thus we increase the null task value 
and cosine similarity threshold ($\alpha$) to 0.26 for \name, Khronos-\thres, and \cg-\thres. 
\begin{table}[h]\scriptsize
    \vspace{1.5mm}
    \setlength{\tabcolsep}{2pt}
    \resizebox{\columnwidth}{!}{%
    {%
    \begin{tabular}{cl ccc ccc ccc}
    \toprule
    &  & \multicolumn{3}{c}{Strict} & \multicolumn{3}{c}{Relaxed} & & & \\
    \cmidrule(r){3-5} \cmidrule(r){6-8}
    Scene & Method & osR$\uparrow$ & osP$\uparrow$ & F1$\uparrow$ & osR$\uparrow$ & osP$\uparrow$ & F1$\uparrow$ & IOU$\uparrow$ & Objs$\downarrow$ & TPF [s]$\downarrow$ \\
    \midrule
    & CG~\cite{Gu24icra-conceptgraphs} & 0.56  & \underline{0.39}  & \underline{0.46}  & \underline{0.89}  & \underline{0.52}  & \underline{0.65}  & 0.06  & 231 & 3.15  \\ 
    & Khronos~\cite{Schmid24rss-khronos} & \textbf{0.83} & 0.16 & 0.27 & 0.83 & 0.17 & 0.28 &  \underline{0.18} & 623 & 1.16\\
    & Clio-Prim & 0.72 & 0.15 & 0.25 & \underline{0.89} & 0.15 & 0.25 & \textbf{0.20} & 956 & 1.14\\
    \rowcolor{blue!20} \cellcolor{white} & CG-\thres & 0.56 & \textbf{0.43} & \textbf{0.49} & \underline{0.89}  & \textbf{0.57} & \textbf{0.70} & 0.06 & 49 & 3.15 \\
    \rowcolor{blue!20} \cellcolor{white}& Khronos-\thres & \textbf{0.83} & 0.19 & 0.31 & 0.83 & 0.20 & 0.32 &  \underline{0.18} & 195 & 1.16\\
    \rowcolor{blue!20} \cellcolor{white}& Clio-\batch & \underline{0.78} & 0.28 & 0.41 & \textbf{0.94} & 0.31 & 0.47 & 0.17 & 96 & 1.16$^*$\\
    \midrule
    \multirow{-9}{*}{\rotatebox{90}{Cubicle}}
    & CG~\cite{Gu24icra-conceptgraphs} & 0.30  & 0.15  & 0.20  & 0.55  & 0.23  & 0.33  & 0.09  & 908 & 12.33 \\
    & Khronos~\cite{Schmid24rss-khronos} & \underline{0.58} &  \underline{0.24} &  \underline{0.34} & \underline{0.61} & 0.25 & 0.35 & \underline{0.13} & 1203 & 1.15\\
    & Clio-Prim & \textbf{0.61} & 0.21 & 0.31 & \underline{0.61} & 0.22 & 0.32 & \textbf{0.16} & 1717 & 1.13\\
    \rowcolor{blue!20}\cellcolor{white}& CG-\thres & 0.27  & 0.19  & 0.22  & 0.55  & \underline{0.29}  & \underline{0.38}  & 0.08  & 247 & 12.33 \\
    \rowcolor{blue!20} \cellcolor{white}& Khronos-\thres & 0.55 &  \underline{0.24} & 0.33 & 0.58 & 0.25 & 0.35 & 0.13 & 351 & 1.15\\
    \rowcolor{blue!20} \cellcolor{white}& Clio-\batch &  \underline{0.58} & \textbf{0.35} & \textbf{0.44} & \textbf{0.76} & \textbf{0.46} & \textbf{0.57} & 0.12 & 224 & 1.15$^*$\\
    \midrule
    \multirow{-9}{*}{\rotatebox{90}{Office}}
    & CG~\cite{Gu24icra-conceptgraphs}  & 0.30 & 0.13 & 0.18 & 0.52 & 0.20 & \underline{0.29} & 0.08 & 908 & 3.54 \\ 
    & Khronos~\cite{Schmid24rss-khronos} & 0.35 & 0.11 & 0.17 & \underline{0.59} & 0.16 & 0.25 & 0.09 & 1081  & 1.03\\
    & Clio-Prim & \textbf{0.48} & 0.12 & 0.19 & \textbf{0.69} & 0.16 & 0.26 & \textbf{0.13} & 1482 & 0.99\\
    \rowcolor{blue!20} \cellcolor{white}& CG-\thres & 0.34 & \textbf{0.23} & \textbf{0.27} & \underline{0.59} & \textbf{0.30} & \textbf{0.40} & 0.08 & 434 & 3.54 \\
    \rowcolor{blue!20} \cellcolor{white}& Khronos-\thres & 0.35  & 0.11 & 0.17  & \underline{0.59} & 0.17 & 0.26 & 0.09  & 363 & 1.03\\
    \rowcolor{blue!20} \cellcolor{white}& Clio-\batch & \underline{0.38} & \underline{0.16} & \underline{0.23} & \textbf{0.69} & \underline{0.28} & \textbf{0.40} & \underline{0.10} & 222 & 1.01$^*$\\
    \midrule
    \multirow{-9}{*}{\rotatebox{90}{Apartment}}
    \end{tabular}}
    } %
    \vspace{-4mm}
    \caption{Results of locating objects of interest via open-set task query for three datasets. 
    We include results for OpenCLIP ViT-H-14. The office, apartment, and cubicle datasets have 
    33, 28, and 18 objects of interest respectively. Results generated with 3090 GPU and Intel i9-12900K. 
    Shaded methods are informed by the list of tasks. First and second-best results are bolded and underlined, respectively.
    $^*$Total time for \name-batch normalized by number of images; 
    clustering step for batch run once on entire graph takes approximately 30 seconds and thus not suitable for online use.}
    \label{table:three_d_results2}
\end{table} 
\subsection{Closed-Set Places Clustering Task List}

For the experiment shown in \cref{table:places}, we report the task prompts used for each scene.
Note that we prefix each categorical prompt with ``an image of \ldots'' to mimic similiar closed-set experiments (\eg{} Replica).

For the Apartment scene, we used
\begin{enumerate}
    \item an image of a kitchen
    \item an image of a bedroom
    \item an image of a doorway
\end{enumerate}

For the Office scene, we used
\begin{enumerate}
    \item an image of a computing workspace
    \item an image of a hallway or corridor
    \item an image of a kitchenette
    \item an image of a conference room
\end{enumerate}

For the Building scene, we used
\begin{enumerate}
    \item an image of a student lounge
    \item an image of a kitchnette or utility closet
    \item an image of a classroom
    \item an image of a conference room
    \item an image of a stairway
    \item an image of a workshop or machine shop
    \item an image of an aircraft hangar of garage
\end{enumerate}

\subsection{Places Clustering Results Visualization}
\label{sec:places_ap_viz}

\begin{figure}[ht]
    \centering
    \begin{subfigure}[ht]{0.47\textwidth}
        \includegraphics[width=\textwidth]{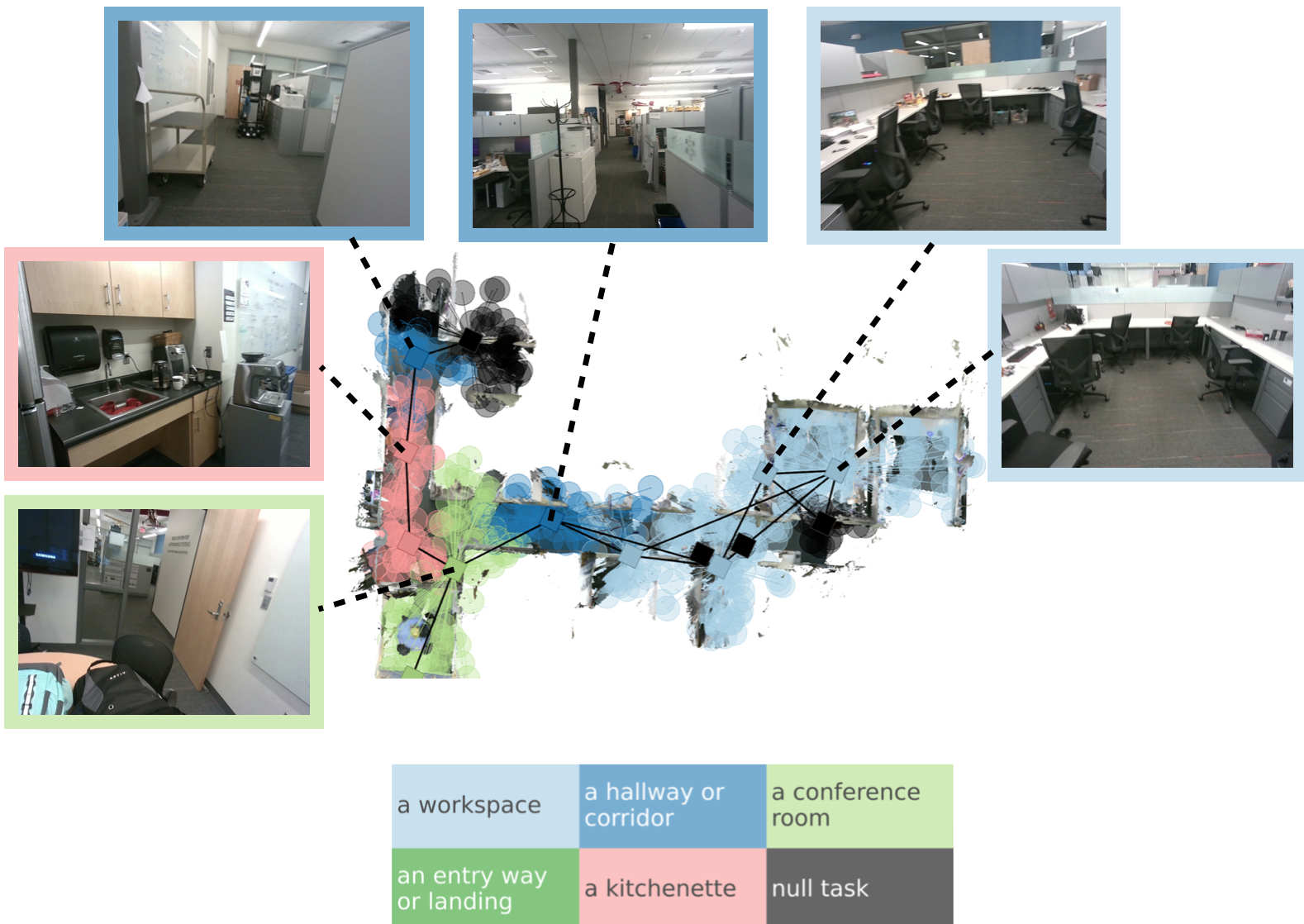}
        \label{fig:places_viz_ap_a}
    \end{subfigure}
    \hspace{\fill}
    \centering
    \begin{subfigure}[ht]{0.47\textwidth}
        \includegraphics[width=\textwidth]{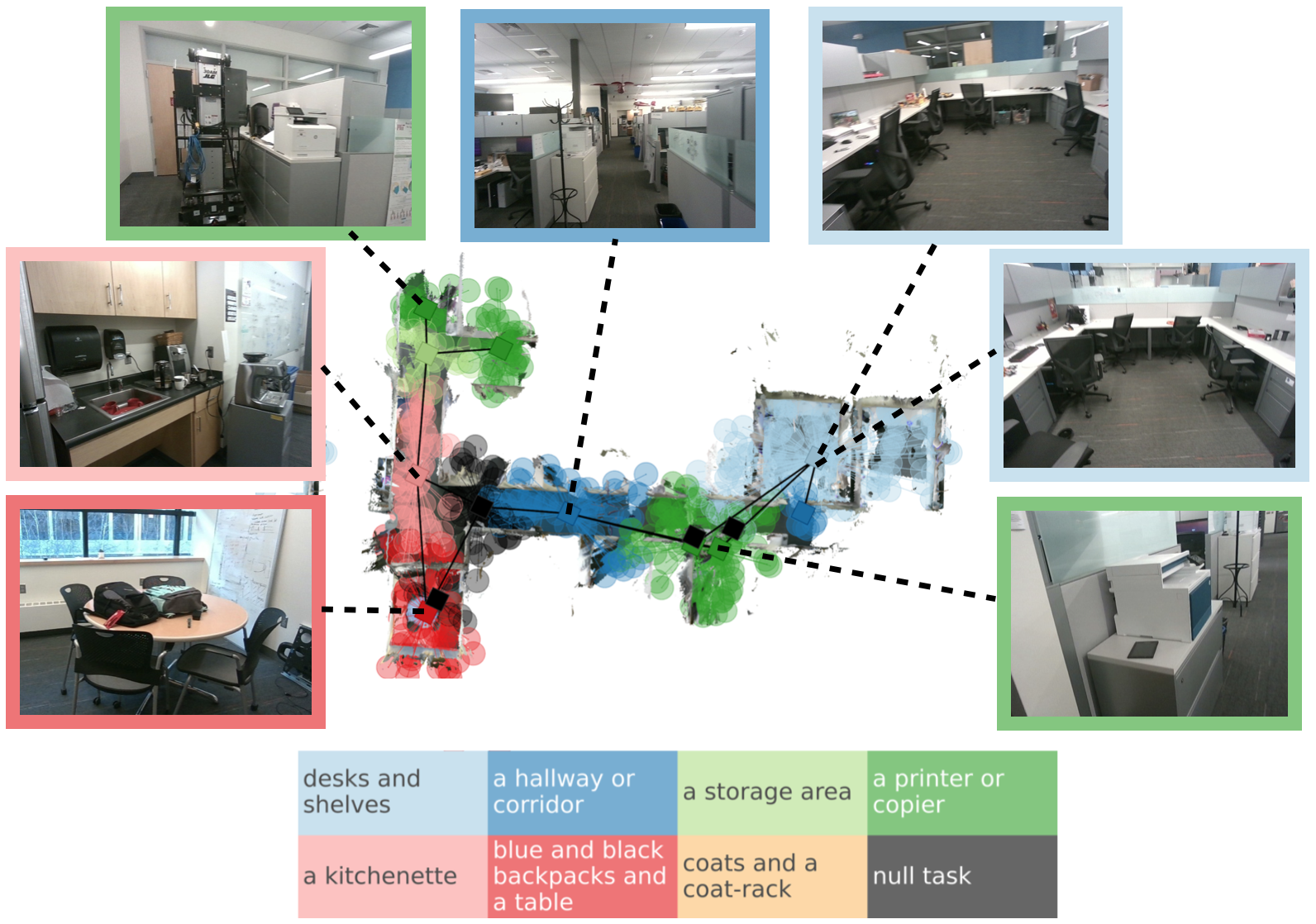}
        \label{fig:places_viz_ap_b}
    \end{subfigure}
    \caption{Visualization of region clustering results on office dataset with example images 
    from regions included for two different task lists.}
    \label{fig:places_viz_ap}
\end{figure}

We include an additional visualization of clustering places into relevant regions on the office dataset by showing
example figures of a subset of the regions in \cref{fig:places_viz_ap} to supporting the meaningfulness of \name's region clustering.

\end{document}